\documentclass[twoside,11pt]{article}

\usepackage{jmlr2e}
\usepackage{mathrsfs}
\usepackage{amsfonts}
\usepackage{amsmath,amssymb}
\usepackage{mathbbold}
\usepackage{color}
\usepackage{algorithm}
\usepackage{algorithmic}

\newcommand{\x}{\mathbf{x}}
\newcommand{\y}{\mathbf{y}}
\newcommand{\ty}{\tilde{\mathbf{y}}}
\newcommand{\f}{\bar{f}}
\newcommand{\D}{\mathcal{D}}
\newcommand{\T}{\mathrm{T}}
\newcommand{\rH}{\mathcal{H}}
\newcommand{\fK}{\mathbf{K}}

\def\bbbe{{\rm I\!E}} 
\def\bbbr{{\rm I\!R}} 
\def\bbbn{{\rm I\!N}} 
\newtheorem{assumption}{Assumption}

\jmlrheading{1}{2000}{1-48}{4/00}{10/00}{Yuyang Wang and Roni Khardon and Pavlos Protopapas}

\ShortHeadings{Grouped Multi-task Learning}{Wang and Khardon and Protopapas}
\firstpageno{1}

\begin{document}

\title{Infinite Shift-invariant Grouped Multi-task Learning for Gaussian Processes{\thanks{This is an extended version of~\citep{wang2010shift}.}}}

\author{\name Yuyang Wang \email ywang02@cs.tufts.edu \\
        \name Roni Khardon \email roni@cs.tufts.edu \\
       \addr Department of Computer Science\\
       Tufts University\\
       Medford, MA 02155, USA
       \AND
       \name Pavlos Protopapas \email pprotopapas@cfa.harvard.edu\\
       \addr Harvard-Smithsonian Center for Astrophysics\\
       Harvard University\\
       Cambridge, MA 02140, USA}

\editor{Unknown}

\maketitle

\begin{abstract}
Multi-task learning leverages shared information among data
 sets to improve the learning performance of individual tasks.
 The paper applies this framework for data where each task is a
 phase-shifted periodic time series.
 In particular, we develop a novel Bayesian nonparametric model
 capturing a mixture of Gaussian processes where each task
 is a sum of a group-specific function and a component capturing
 individual variation, in addition to each task being phase shifted.
 We develop an efficient \textsc{em} algorithm to learn the
 parameters of the model. As a special case we obtain the
 Gaussian mixture model and \textsc{em} algorithm for phased-shifted
 periodic time series.
Furthermore, we extend the proposed model by using a Dirichlet
Process prior and thereby leading to an infinite mixture model that
is capable of doing automatic model selection. A Variational Bayesian
approach is developed for inference in this model.
 Experiments in regression, classification and class discovery
 demonstrate the performance of the proposed models using both
 synthetic data and real-world time series data from
 astrophysics. Our methods are particularly useful when the time
 series are sparsely and non-synchronously sampled.
\end{abstract}


\section{Introduction}
In many real world problems we are interested in learning multiple
tasks while the training set for each task is quite small. For
example, in pharmacological studies, we may be attempting to
predict the concentration of some drug at different times across
multiple patients. Finding a good regression function of an individual
patient based only on his or her measurements can be difficult due to
insufficient training points for the patient. Instead, by using measurements across all the patients, we may be
able to leverage common patterns across patients to obtain better estimates for the population and for each
patient individually.
Multi-task learning captures this intuition aiming to learn
multiple correlated tasks simultaneously. This idea has attracted
much interest in the literature and several approaches have been
applied to a wide range of domains including medical
diagnosis~\citep{biimproved}, recommendation
systems~\citep{dinuzzo2008client} and HIV Therapy
Screening~\citep{bickel2008multi}.
Building on the theoretical framework for single-task learning,
multi-task learning has recently been formulated by~\cite{evgeniou2006learning} as a multi-task regularization
problem in vector-valued Reproducing Kernel Hilbert space.

Several approaches formalizing multi-task learning exist within
Bayesian statistics. Considering hierarchical Bayesian
models~\citep{xue2007multiBB,gelman2004bayesian}, one can view the
parameter sharing of the prior among tasks as a form of multi-task
learning where evidence from all tasks is used to infer the
parameters. Over the past few years, Bayesian models for multi-task
learning were formalized using Gaussian
processes~\citep{yu2005learning,schwaighofer2005lgp,pillonetto2010bayesian}.
In this mixed-effect model, information is shared among tasks by
having each task combine a common (fixed effect) portion and a task
specific portion, each of which is generated by an independent
Gaussian process.

Our work builds on this formulation extending it and the associated
algorithms in several ways.  In particular, we extend the model to
include three new aspects. First, we allow the fixed effect to be
multi-modal so that each task may draw its fixed effect from a
different cluster. Second, we extend the model so that each task may
be an arbitrarily phase-shifted image of the original time series.
This yields our GMT model: the shift-invariant grouped mixed-effect model. 
Alternatively, our model can be viewed as a probabilistic
extension of the Phased K-means algorithm
of~\cite{rebbapragada2009finding} that performs clustering for
phase-shifted time series data and as a non-parametric Bayesian
extension of mixtures of random effects regressions for curve
clustering \citep{gaffney2003curve}. Finally, unlike the existing
models that require the model order to be set a priori, 
our extension in the DP-GMT model uses a
Dirichlet process prior on the mixture proportions so that the number
of mixture components is adaptively determined by the data rather
than being fixed explicitly.

Our main technical contribution is the inference algorithm for the
proposed model. We develop details for the \textsc{em} algorithm for the GMT model and a
Variational \textsc{em} for DP-GMT optimizing the maximum a posteriori (MAP)
estimates for the parameters of the models. Technically, the 
main insights are in estimating the expectation for the coupled
hidden variables (the cluster identities and the task specific
portion of the time series) and in solving the regularized least
squares problem for a set of phase-shifted observations. 
In addition, for the DP-GMT, we show that the variational \textsc{em} algorithm can be implemented with the same complexity as the fixed order GMT without using sampling. Thus the DP-GMT provides an efficient model selection algorithm compared to alternatives such as BIC.
As a special case our algorithm yields the (Infinite) Gaussian mixture model
for phase shifted time series, which may be of independent interest, and
which is a generalization of the algorithms of~\cite{rebbapragada2009finding}
and~\cite{gaffney2003curve}.

Our model primarily captures regression of time series but
because it is a generative model it can be used for class
discovery, clustering, and classification.  We demonstrate the
utility of the model using several
experiments with both synthetic data and real-world time series
data from astrophysics. The experiments show that our model can yield
superior results when compared to the single-task learning and
Gaussian mixture models, especially when each individual task is
sparsely and non-synchronously sampled.
The DP-GMT model yields results that are competitive with model selection
using BIC over the GMT model, at 
much reduced computational cost. 

The remainder of the paper is organized as follows. Section 2 provides an
introduction to the multi-task learning problem and its Bayesian
interpretation and develops the main assumptions of our model.
Section 3 defines the new generative model,
Section 4 develops the \textsc{em} algorithm for it, and the infinite mixture extension is addressed
in Section 5. The experimental results are
reported in Section 6. Related work is discussed in Section 7 and the final
section concludes with a discussion and outlines ideas for future work.

\section{Preliminaries}
Throughout the paper, scalars are denoted using italics, as in
$x,y\in\bbbr$; vectors use bold typeface, as in $\x,\y$, and $x_i$
denotes the $i$th entry of $\x$. For a vector $\x$ and real valued
function $f:\bbbr\rightarrow \bbbr$, we extend the notation for $f$
to vectors so that $f(\x) = [f(x_1),\cdots, f(x_n)]^\T$ where
the superscript $\T$ stands for transposition (and the result is a
column vector). $\mathcal{K}(\cdot,\cdot)$ denotes a kernel function associated to
some reproducing kernel Hilbert space (RKHS) $\mathcal{H}$ and its
norm is denoted as $\|\cdot\|_\mathcal{H}$. To keep the notation simple, $\sum_{j=1}^M$ is
substituted by $\sum_j$ where the index $j$ is not confusing.

\subsection{Multi-task learning with kernel}
Given training set $\mathcal{D}=\{\x_i,y_i\},i=1,\cdots,N$, where
$\x_i=[x_{i1},x_{i2},\cdots,x_{id}]^\T\in
\mathcal{X}\subset\bbbr^d$, single-task learning focuses on finding
a function $f: \mathcal{X}\rightarrow\bbbr$ that best fits and
generalizes the observed data. In the regularization framework,
learning $f$ amounts to solving the following variational
problem~\citep{evgeniou2000rna,scholkopf2002learning}
\begin{equation}
\label{eqn:stl} f^* =
\underset{f\in\mathcal{H}}{\text{argmin}}\left\{ \sum_iV(f(\x_i),
y_i) + \lambda\|f\|^2_\mathcal{H}\right\}
\end{equation}
where $V(\cdot, \cdot)$ is some (typically convex) loss function. The norm $\|\cdot\|_\mathcal{H}$ relates to regularity condition on
the function where a large norm penalizes non-smooth functions. The regularization parameter
$\lambda$ provides a tradeoff between the loss term and the
complexity of the function.

Consider a set of $M$ tasks, with $j$th task $\D^j=(\x_i^j,y_i^j), i
= 1,2,\cdots,n_j$. Multi-task learning seeks to find $f^j$ for each
task simultaneously, which, assuming square loss function, can be formulated as the following
regularization problem
\begin{equation}
\label{eqn:multiK}
  \underset{f^1,\cdots,f^M\in\mathcal{H}}{\text{argmin}}\left\{
  \frac{1}{M}\sum_{j=1}^M\sum_{i=1}^{n_j}(y^j_i - f^j(\x^j_i))^2
  + \lambda\text{PEN}(f^1,f^2,\cdots,f^j)\right\}
\end{equation}
where the penalty term, applying jointly to all the tasks, encodes
our prior information on how smooth the functions are, as well as how
these tasks are correlated with each other. For example, setting the
penalty term to $\sum_j\|f^j\|_\mathcal{H}$ implies that there is
no correlation among the tasks. It further decomposes the optimization
functional to $M$ separate single-task learning problems. On the other hand,
with a shared penalty, the joint regularization can lead to improved
performance. Moreover, we can use a norm in RKHS with a
\emph{multi-task kernel} to incorporate the penalty
term~\citep{micchelli2005learning}. Formally, consider a
vector-valued function $f:\mathcal{X}\rightarrow\bbbr^M$ defined as
$f\triangleq[f^1, f^2, \cdots, f^M]^\T$. Then Equation~(\ref{eqn:multiK})
can be written as
\begin{equation}
  \label{eqn:mtp}
  \underset{f\in\mathcal{H}}{\text{argmin}}\left\{
  \frac{1}{M}\sum_{j=1}^M\sum_{i=1}^{n_j}(y^j_i - f^j(\x^j_i))^2
  + \lambda\|f\|_\mathcal{H}^2\right\}
\end{equation}
where $\|\cdot\|_\mathcal{H}$ is the norm in RKHS with the
multi-task kernel
$\mathcal{Q}:(\mathbf{\Lambda},\mathcal{X})\times(\mathbf{\Lambda},\mathcal{X})\rightarrow
\bbbr$, where $\mathbf{\Lambda}=\{1,2,\cdots,M\}$. As shown by~\cite{evgeniou2006learning}, the \emph{representer theorem} gives the form of the solution to Equation~(\ref{eqn:mtp})
\begin{equation}
  f^\ell(\cdot) = \sum_{j=1}^M\sum_{i=1}^{n_j}c_i^j\mathcal{Q}((\cdot,\ell),(\x_i^j,j))
\end{equation}
with norm
\begin{displaymath}
  \|f\|_{\mathcal{Q}}^2 = \sum_{\ell,k}\sum_{i=1}^{n_\ell}\sum_{j=1}^{n_k}c_i^\ell c_j^k\mathcal{Q}((\x_i^\ell,\ell), (\x_j^k,
  k)).
\end{displaymath}
Let $\mathbf{C} = [{c}_1^1,{c}_2^1,\cdots,
{c}_{n_M}^M]^\T,\mathbf{Y} =
[y^1_1,y^1_2,\cdots,y^M_{n_M}]^\T\in\bbbr^{\sum_jn_j}$ and
$\mathbf{X}=[\x^1_1,\x^1_2,\cdots,\x^M_{n_M}]$, then the
coefficients $\{c_i^j\}$ are given by the following linear system
\begin{equation}
  (\mathbf{Q}+\lambda\mathbb{I})\mathbf{C} = \mathbf{Y}
\end{equation}
where $\mathbf{Q}\in\bbbr^{\sum_jn_j\times \sum_jn_j}$ is the kernel
matrix formed by $\mathbf{X}$.

\subsection{Bayesian formulation}
A Gaussian process is a functional extension for Multivariate Gaussian
distributions. In the Bayesian literature, it has been widely used
in statistical models by substituting a parametric latent
function with a stochastic process with a Gaussian prior
~\citep{rasmussen2006gaussian}. More precisely, under the single-task
setting a simple Gaussian regression model is given by
\begin{displaymath}
  y = f(\x) + \epsilon
\end{displaymath}
where $f$'s prior is a zero mean Gaussian process with covariance
function $\mathcal{K}$ and $\epsilon$ is independent zero mean
white noise with variance $\sigma^2$. Given data set
$\mathcal{D}=\{\x_i,y_i\},i=1,\cdots,N$, let
$\fK=(\mathcal{K}(\x_i,\x_j))_{i,j}$, then $\mathbf{f} \triangleq
[f(\x_1),\cdots, f(\x_N)]^\T\sim\mathcal{N}(\mathbf{0},\fK)$ and the
posterior on $\mathbf{f}$ is given by
\begin{displaymath}
  \Pr(\mathbf{f}|\D) = \mathcal{N}(\fK(\sigma^2\mathbb{I}+\fK)^{-1}\y, \sigma^2(\sigma^2\mathbb{I}+\fK)^{-1}\fK).
\end{displaymath}
The predictive distribution for some test point $\x_*$ distinct from the training examples is
\begin{displaymath}
  \begin{split}
    \Pr(f(\x_*)|\x_*,\D) &= \int \Pr(f(\x_*)|\x_*, f)\Pr(f|\D)df\\
    &= \mathcal{N}\left(\mathbf{k}(\x_*)^\T(\sigma^2\mathbb{I}+\fK)^{-1}\y, \mathcal{K}(\x_*,\x_*)
    -\mathbf{k}(\x_*)^\T(\sigma^2\mathbb{I}+\fK)^{-1}\mathbf{k}(\x_*)\right)
  \end{split}
\end{displaymath}
where $\mathbf{k}(\x_*) =
[\mathcal{K}(\x_1,\x_*),\cdots,\mathcal{K}(\x_N,\x_*)]^\T$. Furthermore,
 under square loss function, the optimizer of Equation~(\ref{eqn:stl})
is equal to the expectation of the predictive
distribution~\citep{rasmussen2006gaussian}. Finally, a Gaussian
process $f$ corresponds to a RKHS $\rH$ with kernel $\mathcal{K}$
such that
\begin{equation}
\label{eqn:RKR}
  \text{cov}[f(\x), f(\y)] = \mathcal{K}(\x,\y)\quad\forall
  \x,\y\in\mathcal{X}.
\end{equation}
In this way, we can express a prior on functions $f$ using 
a zero mean Gaussian process~\citep{lu2008rkh}\footnote{In general, a Gaussian process
  can not be thought of as a distribution on the RKHS, because with
  probability 1, one can find a Gaussian process such that its sample path
  does not belong to the RKHS. However, the equivalence holds between the
  RKHS and the expectation of a Gaussian process conditioned on a finite
  number of observations. For more details on the relationship between RKHS
  and Gaussian processes we refer interested reader to
  ~\cite{seeger2004gaussian}.
}
\begin{equation}
  \label{eqn:fix1}
  \begin{split}
    f &\sim \exp\left\{-\frac{1}{2}\|f\|_\mathcal{H}^2\right\}.\\
  \end{split}
\end{equation}
Applying this framework in the context of multi-task learning, the model is
given by
\begin{displaymath}
  \y^j_i = f^j(\x^j_i) + \epsilon_{ij}
\end{displaymath}
where $f^j$ are zero mean Gaussian processes and $\epsilon_{ij}$
captures i.i.d.\ zero-mean noise with variance $\sigma^2$. 
\cite{pillonetto2010bayesian} formalize the connection
between \emph{multi-task kernel} $\mathcal{Q}$ and covariance
function among $\{f^j\}$ using
\begin{equation}
\label{eqn:mcov}
  \text{cov}[f^i(\x), f^j(\x')] = \mathcal{Q}((\x, i), (\x', j)),\quad i,j=1,\cdots,M.
\end{equation}

\subsection{Basic Model Assumptions}
\label{sec:shift} Given data $\{\D^j\}$, the so-called nonparametric Bayesian mixed-effect
model~\citep{lu2008rkh,pillonetto2010bayesian} captures each task
$f^j$ with respect to $\D^j$ using a sum of an average effect
function and an individual variation for each specific task,
\begin{displaymath}
  f^j(x) = \bar{f}(x) + \tilde{f}^j(x), \quad j=1,\cdots, M.
\end{displaymath}
This assumes that the fixed-effect (mean function) $\f$ is
sufficient to capture the behavior of the data, an assumption that
is problematic for distributions with several modes. To address
this, we introduce a mixture model allowing for multiple modes (just
like standard Gaussian mixture model (GMM)), but maintaining the
formulation using Gaussian processes. This amounts to adding a group
effect structure and leads to the following assumption:
\begin{assumption}
  For each $j$ and $\x\in\mathcal{X}$,
  \begin{equation}
    f^j(\x) = \bar{f}_{z_j}(\x) + \tilde{f}^j(\x), \quad j=1,\cdots, M
  \end{equation}
  where $\{\bar{f}_s\},s=1,\cdots,k$ and $\tilde{f}^j$ are zero-mean
  Gaussian processes and $z_j\in\{1,\cdots,k\}$.
  In addition, $\{\bar{f}_s\}$ and $\tilde{f}^j$ are assumed to be mutually independent.
\end{assumption}
With the grouped-effect model and groups predefined, one can define a
kernel that relates (with non zero similarity) only points from the
same example or points for different examples but the same center as
follows
\begin{displaymath}
  \mathcal{Q}((\x,i), (\x',j)) =
  \delta_{z_i,z_j}\overline{\mathcal{K}}_{z_i}(\x,\x') +
   \delta_{i,j}\widetilde{\mathcal{K}}_i(\x,\x')
\end{displaymath}
where
\begin{displaymath}
  \begin{cases}
    \overline{\mathcal{K}}_{z_i}(\x,\x') = \text{cov}[\bar{f}_{z_i}(\x), \bar{f}_{z_i}(\x')],\\
    \widetilde{\mathcal{K}}_i(\x,\x') = \text{cov}[\tilde{f}^j(\x),
    \tilde{f}^j(\x')].
  \end{cases}
\end{displaymath}
However, in our work the groups are not known in advance and we cannot use
this formulation. Instead we use a single kernel to relate all tasks.

The second extension allows us to handle phase shifted time series.
In some applications, we face the challenge of learning a periodic
function $f^j:\bbbr\rightarrow\bbbr$ on a single period $T$ from
samples $\D = \{\x^j,\y^j\},\x^j,\y^j\in\bbbr^{n_j}, j=1,\cdots,M$,
where similar functions in a group
differ only in their phase. In the following assumption, the model
of primary focus in this paper is presented, which extends the
mixed-effect model to capture both shift-invariance and the
clustering property.
\begin{assumption}
  For each $j$ and $x\in[0,T)$,
  \begin{equation}
    f^j(x) = [\bar{f}_{z_j}*\delta_{t_j}](x) + \tilde{f}^j(x), \quad, j=1,\cdots, M
  \end{equation}
  where $z_j\in\{1,\cdots,k\}$, $\{\bar{f}_s\},s=1,\cdots,k$ and $\tilde{f}^j$ are
  zero-mean Gaussian processes, $*$ stands for circular convolution and
  $\delta_{t_j}$ is the Dirac $\delta$ function with support at $t_j\in[0,T)$.\footnote{
  Given a periodic function
  $f$ with period $T$, its circular convolution with another function $h$ is defined as
  \begin{displaymath}
    (f*h)(t) \triangleq \int_{t_0}^{t_0+T}f(t-\tau)h(\tau)d\tau
  \end{displaymath}
  where $t_0$ is arbitrary in $\bbbr$ and $f*h$ is also a periodic function with period $T$. Using the definition we see that,
  \begin{displaymath}
    f*\delta_{t_j}(t) = f( t - t_j),
  \end{displaymath}
and thus $*$ performs a right shift of $f$ or in other words performs a phase shift of $t_j$ on $f$.}
  In addition, $\{\bar{f}_s\},\tilde{f}^j$ are assumed to be mutually independent.
\end{assumption}

\section{Shift-invariant Grouped mixed-effect model}

\begin{figure}
\centering
        \includegraphics[width=5cm]{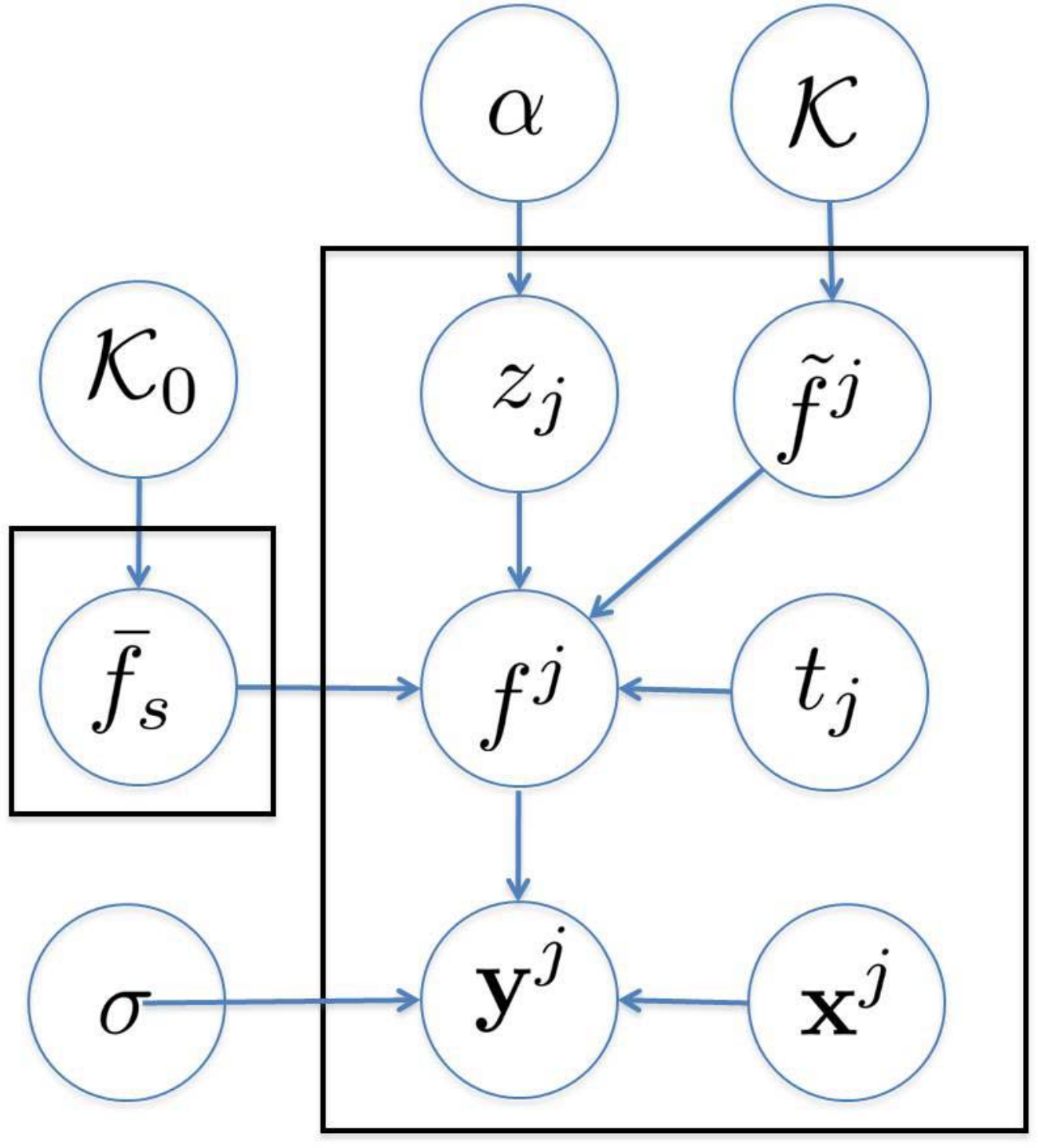}
        \caption{GMT: Plate graph}
        \label{fig:gmt_graph}
\end{figure}

In Assumption 1, if we know the cluster assignment of each task, then the
model decomposes to $k$ mixed-effect models which is the case investigated
in~\citep{pillonetto2010bayesian,evgeniou2006learning}. Similar results can
be obtained for Assumption 2.  However, prior knowledge of cluster membership is
often not realistic.  In this section, based on Assumption 2, a probabilistic
generative model is formulated
to capture the case of unknown clusters.
 We start by formally defining the generative model,
which we call \emph{Shift-invariant Grouped mixed-effect Model}
(GMT). In this model, $k$ group effect functions are assumed to
share the same Gaussian prior characterized by $\mathcal{K}_0$. The
individual effect functions are Gaussian processes with covariance
function $\mathcal{K}$. The model is shown in
Figure~\ref{fig:gmt_graph} and it is characterized by parameter set
$\mathcal{M} = \{\mathcal{K}_0,\mathcal{K},\boldsymbol\alpha, \{t_j\}, \sigma^2\}$
and summarized as follows
\begin{enumerate}
  \item Draw $\f_s|\mathcal{K}_0 \sim \exp\left\{-\frac{1}{2}\|\f_s\|_{\mathcal{H}_0}^2\right\},\quad s=1,2,\cdots,
  k$
  \item For the $j$th time series
  \begin{itemize}
    \item Draw $z_j|\mathbf{\boldsymbol\alpha} \sim
    \text{Discrete}(\mathbf{\boldsymbol\alpha})$
    \item Draw $\tilde{f}^j|\mathcal{K} \sim \exp\left\{-\frac{1}{2}\|\tilde{f}^j\|_\mathcal{H}^2\right\}$
    \item Draw $\y^j | z_j, f^j, \x^j, t_j, \sigma^2 \sim \mathcal{N}\left(f^j(\x^j),
    \sigma^2\mathbb{I}\right)$, where $f^j = \f_{z_{j}}*\delta_{t_j}+\tilde{f}^j$
  \end{itemize}
\end{enumerate}
where $\boldsymbol\alpha$ is the mixture proportion. Additionally,
denote $\mathcal{X}=\{\x^1, \x^2, \cdots,\x^M\}$ and $\mathcal{Y}=
\{\y^1, \y^2, \cdots,\y^M\}$, where $\x^j$ are the time points when
each time series is sampled and $\y^j$ are the corresponding
observations.

We assume that the group effect kernel $\mathcal{K}_0$ and the number of
centers $k$ are known. The assumption on $\mathcal{K}_0$ is reasonable, in
that normally we can get more information on the shape of the mean waveforms,
thereby making it possible to design kernel for $\mathcal{H}_0$. On the other hand, the
individual variations are more arbitrary and therefore $\mathcal{K}$ is not
assumed to be known.
The assumption that $k$ is known requires some form of model selection.
An extension using a non-parametric
Bayesian model, the \emph{Dirichlet process}~\citep{Teh2010a},
that does not limit
$k$ is discussed in the section~\ref{sec:infinite}. The group effect
$\{\f_s\}$, individual shifts $\{t_j\}$, noise variance $\sigma^2$ and the
kernel for individual variations $\mathcal{K}$ are unknown and need to be
estimated.
The cluster assignments $\{z_j\}$ and individual variation $\{\tilde{f}^j\}$ are
treated as hidden variables. Note that one could treat
$\{\f_s\}$ too as hidden variables, but we prefer to get a concrete estimate
for these variables because of their role as the mean waveforms in our model.

The model above is a standard model for regression. We propose to use it for classification by learning a mixture model for each class and using the \emph{Maximum A Posteriori} (MAP) probability for the class for classification. In particular, consider a training set that has $L$ classes, where the $j$th instance is given by $\D^j=(\x^j,\y^j,o^j)\in \bbbr^{n_j}\times \bbbr^{n_j}\times\{1,2,\cdots,L\}$. Each observation $(\x^j, \y^j)$ is
   given a label from $\{1,2,\cdots,L\}$. The problem is to learn the model $\mathrm{M}_\ell$ for each class ($L$ in total)
   separately and the classification rule for a new instance $(\x,\y)$ is given by
   \begin{equation}
   \label{eqn:class}
      o =
      \underset{\ell=\{1,\cdots,L\}}{\text{argmax}}\Pr(\y|\x; \mathrm{M}_\ell)\Pr(\ell).
   \end{equation}
As we show in our experiments, the generative model can provide
explanatory power for the application while giving excellent
classification performance.

\section{Parameter Estimation}

 Given data set $\D=\{\x^j,\y^j\}=\{x_i^j,y_i^j\}, i=1,\cdots,n_j, j=1,\cdots,M$, the learning process aims to find the MAP estimates of the parameter set $\mathcal{M}=\{\boldsymbol\alpha, \{\f_s\}, \{t_j\}, \sigma^2, \mathcal{K}\}$
\begin{equation}
 \label{eqn:EM}
        \mathcal{M}^* = \underset{\mathcal{M}}{\text{argmax}}
        \left(\Pr(\mathcal{Y}|\mathcal{X}; \mathcal{M})\times \Pr[\{\f_s\};\mathcal{K}_0]\right).
\end{equation}
 The direct optimization of Equation~(\ref{eqn:EM}) is analytically intractable because of coupled sums that come
 from the mixture distribution.  To solve this problem, we resort to the \textsc{em} algorithm \citep{dlr-midea-77}.
 The \textsc{em} algorithm is an iterative
method for optimizing the maximum likelihood (ML) or MAP
estimates of the parameters in the context of hidden variables.
In our case, the hidden variables are
$\mathbf{z}=\{z_{j}\}$ (which is the same as in standard GMM), and
$\mathbf{f} = \{\mathbf{f}_j\triangleq\tilde{f}^j(\x^j)\},
j=1,\cdots,M$. The algorithm iterates between the following
expectation and maximization
steps until it converges to a local
maximum.

\subsection{Expectation step}
In the \textbf{E}-step, we calculate
  \begin{equation}
  \label{eqn:eme}
    Q(\mathcal{M},\mathcal{M}^g) = \bbbe_{\{\mathbf{z},\mathbf{f}|\mathcal{X,Y};
    \mathcal{M}^g\}}\left[\log\left\{\Pr(\mathcal{Y},\mathbf{f,z}|\mathcal{X}; \mathcal{M})\times \Pr[\{\f_s\};\mathcal{K}_0]\right\}\right]
  \end{equation}
  where $\mathcal{M}^g$ stands for estimated parameters from the last iteration.
  For our model, the difficulty comes from estimating the expectation with
  respect to the coupled latent variables $\{\mathbf{z}, \mathbf{f}\}$. In the
  following, we show how this can be done. First notice that,
  \begin{displaymath}
    \Pr(\mathbf{z}, \mathbf{f}|\mathcal{X,Y}; \mathcal{M}^g) = \prod_j\Pr(z_j, \mathbf{f}_j|\mathcal{X,Y}; \mathcal{M}^g)
  \end{displaymath}
  and further that
  \begin{equation}
  \label{eqn:zj}
    \Pr(z_j, \mathbf{f}_j|\mathcal{X,Y}; \mathcal{M}^g) = \Pr(z_j|\x^j, \y^j; \mathcal{M}^g)\times\Pr(\mathbf{f}_j|z_j, \x^j, \y^j; \mathcal{M}^g).
  \end{equation}
  The first term in Equation~(\ref{eqn:zj}) can be further written as
  \begin{equation}
  \label{eqn:mum}
  \begin{split}
    \Pr(z_j|\x^j, \y^j; \mathcal{M}^g) &\propto \Pr(z_j; \mathcal{M}^g)\Pr(\y^j|z_j, \x^j; \mathcal{M}^g)\\
    &= \Pr(z_j; \mathcal{M}^g)\int \Pr(\y^j, \mathbf{f}_j|z_j, \x^j; \mathcal{M}^g)d\mathbf{f}_j\\
    &= \Pr(z_j; \mathcal{M}^g)\int \Pr(\y^j|\mathbf{f}_j, z_j, \x^j; \mathcal{M}^g)\Pr(\mathbf{f}_j;\mathcal{M}^g)d\mathbf{f}_j
  \end{split}
  \end{equation}
  where $\Pr(z_j; \mathcal{M}^g)$ is specified by the parameters estimated
  from last iteration. Since $z_j$ is given, the second term is the marginal
  distribution that can be calculated using
a Gaussian process regression model.
In particular, denoting $\mathbf{\f}^j = \f_{z_j}*\delta_{t_j}(\x^j)$
we get
  \begin{displaymath}
    \begin{bmatrix}
      \y^j\\
      \mathbf{f}^j
    \end{bmatrix}\sim
    \mathcal{N}\left(
    \begin{bmatrix}
      \mathbf{\f}^j\\
      0
    \end{bmatrix},
    \begin{bmatrix}
      \mathbf{K}^g_j + \sigma^2\mathbb{I} & \quad\mathbf{K}^g_j\\
      \mathbf{K}^g_j & \quad\mathbf{K}^g_j
    \end{bmatrix}\right)
  \end{displaymath}
  where $\mathbf{K}^g_j$ is the kernel matrix for the $j$th task using parameters from last iteration, i.e. $\mathbf{K}^g_j = (\mathcal{K}(x_{i}^j,x_{l}^j))_{il}$. Therefore,  the marginal distribution is
  \begin{equation}
   \label{eqn:14p}
    \y^j|z_j \sim\mathcal{N}(\mathbf{\f}^j, \mathbf{K}^g_j + \sigma^2\mathbb{I}).
  \end{equation}
  Next consider the second term in Equation~(\ref{eqn:zj}). Given $z_j$, we know that $f^j=\f_{z_j}+\tilde{f}^j$, i.e. there is no uncertainty about the identity of $\f_{z_j}$ and therefore the calculation amounts to estimating the posterior distribution under
  standard Gaussian process regression, that is
  \begin{displaymath}
      \begin{split}
        &\y^j-\mathbf{\f}^j \sim \mathcal{N}(\tilde{f}^j(\x^j), \sigma^2\mathbb{I})\\
        &\qquad \tilde{f}^j \sim \exp\left\{-\frac{1}{2}\|\tilde{f}^j\|_\mathcal{K}^2\right\}\\
      \end{split}
  \end{displaymath}
  and the conditional distribution is given by
  \begin{equation}
  \label{eqn:fjconditional}
    \mathbf{f}_j|z_j,\x^j,\y^j \sim \mathcal{N}(\mu^g_{j}, \mathbf{C}^g_{j})
  \end{equation}
  where $\mathbf{\mu}^g_{j}$ is the posterior mean
  \begin{equation}
  \label{eqn:mujs}
      \mathbf{\mu}^g_{j} = \mathbf{K}^g_j(\mathbf{K}^g_j + \sigma^2\mathbb{I})^{-1}(\y^j - \mathbf{\f}^j)
    \end{equation}
    and $\mathbf{C}_{j}^g$ is the posterior covariance of $\mathbf{f}_j$
    \begin{equation}
    \label{eqn:cjs}
      \mathbf{C}_{j}^g =\mathbf{K}^g_j-\mathbf{K}^g_j(\mathbf{K}^g_j + \sigma^2\mathbb{I})^{-1}\mathbf{K}^g_j.
    \end{equation}
  Since Equation~(\ref{eqn:mum}) is multinomial and $\mathbf{f}_j$ is Normal
in (\ref{eqn:fjconditional}), the marginal distribution of $\mathbf{f}_j$ is a Gaussian mixture distribution given by \begin{displaymath}
    \begin{split}
    \Pr(\mathbf{f}_j|\x^j, \y^j; \mathcal{M}^g) &= \sum_s\Pr(z_j=s|\x^j, \y^j; \mathcal{M}^g)\\
    &\qquad\qquad\times
     \mathcal{N}\left(\mu_j,\mathbf{C}_j|z_j=s;\mathcal{M}^g\right), \quad s=1,\cdots,k.
  \end{split}
  \end{displaymath}
  To work out the concrete form of $Q(\mathcal{M},\mathcal{M}^g)$, denote $z_{il}=1$ iff $z_i = l$. Then the complete data likelihood can be reformulated as
    \begin{displaymath}
    \begin{split}
      \mathcal{L} &= \Pr(\mathcal{Y},\mathbf{f,z}; \mathcal{X,M})\\
      &= \prod_j\prod_s
      \left[\alpha_s\Pr(\y^j,\mathbf{f}_j|z_j=s;\mathcal{M})\right]^{z_{js}}\\
      &= \prod_j\prod_s
      \left[\alpha_s\Pr(\y^j|\mathbf{f}_j,z_j=s;\mathcal{M})\Pr(\mathbf{f}_j;\mathcal{M})\right]^{z_{js}}\\
    \end{split}
    \end{displaymath}
    where we have used the fact that exactly one $z_{js}$ is 1 for each $j$
    and included the last term inside the product over $s$ for
    convenience. Then Equation~(\ref{eqn:eme}) can be written as
    \begin{displaymath}
        \begin{split}
      Q(\mathcal{M},\mathcal{M}^g) &= -\frac{1}{2}\sum_s\|f_s\|_{\mathcal{H}_0}^2 +
      \bbbe_{\{\mathbf{z,f}|\mathcal{X,Y;M}^g\}}\left[\log\mathcal{L}\right].
        \end{split}
    \end{displaymath}
    Denote the second term by $\widetilde{Q}$. By a version of Fubini's theorem~\citep{stein2005real} we have
    \begin{equation}
      \label{eqn:tQ}
      \begin{split}
        \widetilde{Q} &= \bbbe_{\{\mathbf{z}|\mathcal{X,Y;M}^g\}}\bbbe_{\{\mathbf{f}|\mathbf{z},
      \mathcal{X,Y;M}^g\}}\left[\log\mathcal{L}\right]\\
        &= \sum_{\mathbf{z}}\Pr(\mathbf{z}|\mathcal{X,Y;M}^g)\Bigg\{\sum_j\sum_sz_{js}\\
        &\qquad \times\int d\Pr(\mathbf{f}_j|z_j=s)\log\left[\alpha_s\Pr(\y^j|\mathbf{f}_j,z_j=s;\mathcal{M})
        \Pr(\mathbf{f}_j;\mathcal{M})\right]\Bigg\}.\\
      \end{split}
    \end{equation}
    Now because the last term in Equation~(\ref{eqn:tQ}) does not include any $z_i$, the equation can be further decomposed as
    \begin{equation}
    \label{eqn:18p}
      \begin{split}
        \widetilde{Q}&= \sum_j\sum_s\left(\sum_{\mathbf{z}}\Pr(\mathbf{z}|\mathcal{X,Y;M}^g)z_{js}\right)\\
        &\qquad\times\left\{\int d\Pr(\mathbf{f}_j|z_j=s)\log[\alpha_s\Pr(\y^j|\mathbf{f}_j,z_j=s;\mathcal{M})\Pr(\mathbf{f}_j;\mathcal{M})]
        \right\}\\
        &= \sum_j\sum_s\gamma_{js}\int d\Pr(\mathbf{f}_j|z_j=s)\log\left[\alpha_s\Pr(\y^j|\mathbf{f}_j,z_j=s;\mathcal{M})
        \Pr(\mathbf{f}_j;\mathcal{M})\right]\\
        &= \sum_j\sum_s\gamma_{js}\bbbe_{\{\mathbf{f}_j|z_j=s, \x^j, \y^j; \mathcal{M}^g\}}\left[\log\alpha_s + \log\left(\Pr(\y^j|\mathbf{f}_j,z_j=s;\mathcal{M})\right) +
        \log\left(\Pr(\mathbf{f}_j;\mathcal{M})\right)\right]
      \end{split}
    \end{equation}
    where
    \begin{equation}
    \label{eqn:gammajs}
      \gamma_{js} = \bbbe[z_{js}|\y^j, \x^j;\mathcal{M}^g] =
      \frac{\Pr({z}_j=s|\x^j,\y^j;\mathcal{M}^g)}{\sum_s\Pr({z}_j=s|\x^j,\y^j;\mathcal{M}^g)}
    \end{equation}
    can be calculated from Equation~(\ref{eqn:mum}) and (\ref{eqn:14p}) and $\gamma_{js}$ can be viewed as a fractional label indicating how likely
    the $j$th task is to belong to the $s$th group. Recall that
    $\Pr(\y^j|\mathbf{f}_j,z_j=s)$ is a normal distribution given by
$\mathcal{N}([\f_{z_j}*\delta_{t_j}](\x^j) + \mathbf{f}_j, \sigma^2\mathbb{I})$
and $\Pr(\mathbf{f}_j;\mathcal{M})$ is a standard multivariate Gaussian distribution determined by its prior
  \begin{displaymath}
      \Pr(\mathbf{f}_j;\mathcal{M})=\frac{1}{\sqrt{(2\pi)^{n_j}|\mathbf{K}_j|}}
      \exp\left\{-\frac{1}{2}\mathbf{f}_j^T\mathbf{K}_j^{-1}\mathbf{f}_j\right\}.
    \end{displaymath}
    Using these facts and Equation~(\ref{eqn:18p}),  $Q(\mathcal{M},\mathcal{M}^g)$ can be re-formulated as


  \begin{equation}
  \begin{split}
    Q(\mathcal{M},\mathcal{M}^g) &=
    -\frac{1}{2}\sum_s\|\f_s\|_{\mathcal{H}_0}^2 - \sum_jn_j\log\sigma + \sum_j\sum_s\gamma_{js}\log\alpha_s \\
    &\qquad - \frac{1}{2\sigma^2}\sum_j\sum_s\gamma_{js} \bbbe_{\{\mathbf{f}_j|z_j=s, \x^j, \y^j;
     \mathcal{M}^g\}}\left[\|\y^j - [\f_s*\delta_{t_j}](\x^j) - \mathbf{f}_j\|^2\right]\\
    &\qquad
-\frac{1}{2} \sum_j\log|\mathbf{K}_j| - \frac{1}{2}\sum_j\sum_s\gamma_{js}
\bbbe_{\{\mathbf{f}_j|z_j=s, \x^j, \y^j;
    \mathcal{M}^g\}}\left(\mathbf{f}^\T_j\mathbf{K}^{-1}_j\mathbf{f}_j\right)
  \end{split}
  \end{equation}

We next develop explicit closed forms for the remaining expectations. For the
first, note that for $\x\sim\mathcal{N}(\mu, \mathbf{\Sigma})$ and a constant
vector $\mathbf{a}$,
    \begin{displaymath}
      \begin{split}
        \bbbe[\|\mathbf{a} - \x\|^2] &= \bbbe[\|\mathbf{a}\|^2 - 2\langle\mathbf{a}, \x\rangle + \|\x\|^2]\\
        &= \|\mathbf{a}\|^2 - 2\langle\mathbf{a}, \bbbe[\x]\rangle + \bbbe[\x]^2 + \text{Tr}(\mathbf{\Sigma})\\
        &= \|\mathbf{a} - \mu\|^2 + \text{Tr}(\mathbf{\Sigma}).
      \end{split}
    \end{displaymath}
Therefore the expectation is
  \begin{equation}
  \label{eqn:exp1}
  \begin{split}
\bbbe_{\{\mathbf{f}_j|z_j=s, \x^j, \y^j;
     \mathcal{M}^g\}}
&
\left[\|\y^j - [\f_s*\delta_{t_j}](\x^j) - \mathbf{f}_j\|^2\right]
= \frac{1}{2\sigma^2}\sum_j\text{Tr}(\mathbf{C}_j^g) \\
    &\qquad +\frac{1}{2\sigma^2}\sum_j\sum_s\gamma_{js}
    \left(\|\y^j - [f_s*\delta_{t_j}](\x^j) - \mu_{js}\|^2\right) \\
  \end{split}
  \end{equation}
  where $\mu_{js} = \bbbe_{\{\mathbf{f}_j|z_j=s, \x^j, \y^j;
    \mathcal{M}^g\}}[\mathbf{f}_{j}]$ is as in Equation~(\ref{eqn:mujs}) where we
  set $z_j = s$ explicitly.

For the second expectation we have
\begin{displaymath}
      \begin{split}
        \bbbe_{\{\mathbf{f}_j|z_j=s, \x^j, \y^j;
    \mathcal{M}^g\}}\left(\mathbf{f}^\T_j\mathbf{K}^{-1}_j\mathbf{f}_j\right) &= \bbbe_{\{\mathbf{f}_j|z_j=s, \x^j, \y^j;
         \mathcal{M}^g\}}\left[\text{Tr}\left(\mathbf{f}^\T_j\mathbf{K}^{-1}_j\mathbf{f}_j\right)\right]\\
         &= \bbbe_{\{\mathbf{f}_j|z_j=s, \x^j, \y^j;          \mathcal{M}^g\}}\left[\text{Tr}\left(\mathbf{K}^{-1}_j\mathbf{f}_j\mathbf{f}^\T_j\right)\right]\\
         &= \text{Tr}\left(\bbbe_{\{\mathbf{f}_j|z_j=s, \x^j, \y^j;
         \mathcal{M}^g\}}[\mathbf{K}^{-1}_j\mathbf{f}_j\mathbf{f}^\T_j]\right)\\
         &= \text{Tr}\left(
        \mathbf{K}_j^{-1}(\mathbf{C}_{j}^g + \mu_{js}^g(\mu_{js}^g)^\T)\right).
      \end{split}
    \end{displaymath}

  \subsection{M-step}
  In this step, we aim to find
  \begin{equation}
      \begin{split}
        \mathcal{M}^* &= \underset{\mathcal{M}}{\text{argmax}}\ Q(\mathcal{M},\mathcal{M}^g)
      \end{split}
  \end{equation}
  and use $\mathcal{M}^*$ to update the model parameters. Using the results
  above
  this can be decomposed into three separate optimization problems as follows:
  \begin{displaymath}
      \begin{split}
    \mathcal{M}^*
        &= \underset{\mathcal{M}}{\text{argmax}}\ Q_1((\{\f_s\}, \{\delta_{t_j}\}, \sigma)) \\
        &\qquad + Q_2(\mathcal{K}) +
        \left\{\sum_j\sum_s\gamma_{js}\log\alpha_s\right\}.
        \end{split}
  \end{displaymath}
  That is, $\boldsymbol\alpha$ can be estimated easily using its separate
  term, $Q_1$ is only a function of $(\{f_s\}, \{{t_j}\}, \sigma)$ and $Q_2$
  depends only on $\mathcal{K}$, and we have
  \begin{equation}
  \label{eqn:q1final}
  \begin{split}
       Q_1(\{\f_s\},\{t_j\},\sigma^2) &= \frac{1}{2}\sum_s\|\f_s\|_{\mathcal{K}_0}^2 + \sum_jn_j\log\sigma
       + \frac{1}{2\sigma^2}\sum_j\text{Tr}(\mathbf{C}_j^g) \\
    &\qquad +\frac{1}{2\sigma^2}\sum_j\sum_s\gamma_{js}
    \left(\|\y^j - [f_s*\delta_{t_j}](\x^j) - \mu_{js}\|^2\right) \\
  \end{split}
  \end{equation}
%
and
    \begin{equation}
    \label{eqn:optK}
      \begin{split}
        Q_2(\mathcal{K})
          &= -\frac{1}{2} \sum_j\log|\mathbf{K}_j| - \frac{1}{2}\sum_j\sum_s\gamma_{js}
        \text{Tr}\left(
        \mathbf{K}_j^{-1}(\mathbf{C}_{j}^g + \mu_{js}^g(\mu_{js}^g)^\T)\right).\\
      \end{split}
    \end{equation}
  The optimizations for $Q_1$ and $Q_2$ are described separately in the following two subsections.

\subsubsection{Learning $\{\f_s\},\{t_j\},\sigma^2$}


To optimize Equation~(\ref{eqn:q1final}) we assume first that $\sigma$ is given. In
this case,
optimizing $\{\f_s\}, \{t_j\}$ decouples into $k$ sub-problems,
finding $s$th group effect $\f_s$ and its corresponding shift $\{t_j\}$.
Denoting the residual $\ty^j=\y^j-\mu_{js}$,
where $\mu_{js} = \bbbe[\mathbf{f}_j|\y^j,z_j = s]$,
the problem becomes
 \begin{equation}
 \label{eqn:prob}
   \underset{f\in\mathcal{H}_0, t_1,\cdots, t_M\in[0,T)}{\text{argmin}}
   \left\{\frac{1}{2\sigma^2}\sum_j\gamma_{js}\sum_{i=1}^{n_j}
   (\ty^j_i - [f*\delta_{t_j}](\x^j_i))^2 + \frac{1}{2}\|f\|^2_{\mathcal{H}_0}\right\}.
 \end{equation}
 Note that different $\x^j,\y^j$ have different dimensions $n_j$ and they are
 not assumed to be sampled at regular intervals. For further development, following ~\cite{pillonetto2010bayesian}, it is useful to
 introduce the distinct vector  $\breve{\x}\in\bbbr^{\bbbn}$ whose component are the
 distinct elements of $\mathcal{X}$. For example if $\x^1 = [1, 2, 3]^T,
 \x^2 = [2, 3, 4, 5]^T$, then $\breve{\x} = [1,2,3,4,5]^T$. For $j$th
 task, let the binary matrix $C^k$ be such that
 \begin{displaymath}
  \x^j = C^j\cdot \breve{\x}, \quad f(\x^j) = C^j\cdot f(\breve{\x}).
 \end{displaymath}
 That is, $C^j$ extracts the values corresponding to the $j$th task
from the full vector.
If $\{t_j\}$ are fixed, then the optimization in Equation~(\ref{eqn:prob}) is standard and the representer theorem gives the form of the solution as
\begin{equation}
\label{eqn:solf}
  f(\cdot) = \sum_{i=1}^{\bbbn}c_i\mathcal{K}_0(\breve{x}_i, \cdot).
\end{equation}
Denoting the kernel matrix as
$\mathfrak{K}=\mathcal{K}_0(\breve{x}_i, \breve{x}_j),i,j=1,\cdots,\bbbn$
, $\mathbf{c} = [c_1,\cdots, c_\bbbn]^T$ and we get $f(\breve{\x}) = \mathfrak{K}\mathbf{c}$.
 To
simplify the optimization we assume that $\{t_j\}$ can only take
values in the discrete space $\{\tilde{t}_1,\cdots,\tilde{t}_L\}$,
that is, $t_j = \tilde{t}_i$, for some $i\in 1,2,\cdots,L$ (e.g.,\ a
fixed finite fine grid), where we always choose $\tilde{t}_1 = 0$.
Therefore, we can write $\left[f*\delta_{t_j}\right](\breve{\x}) =
\widetilde{\mathcal{K}}^{\T}_{t_j}\mathbf{c}$, where
$\widetilde{\mathcal{K}}_{t_j}$ is $\mathcal{K}_0(\breve{\x},[
(\breve{\x} - \tilde{t}_j)\mod T])$. Accordingly, Equation~(\ref{eqn:prob}) is reduced to
\begin{equation}
    \label{eqn:probls}
    \underset{\mathbf{c}\in\bbbr^\bbbn, t_1,\cdots,
    t_j\in\{\tilde{t}_i\}}{\text{argmin}}\left\{
    \sum_j\gamma_{js}\|\ty^j-C^j\cdot\widetilde{\mathcal{K}}^\T_{t_j}\mathbf{c}\|^2 + \frac{1}{2} \mathbf{c}^\T\mathfrak{K}\mathbf{c}\right\}.
\end{equation}
To solve this optimization, we follow a cyclic optimization approach where we alternate between steps of optimizing $f$ and $\{t_j\}$ respectively,
\begin{itemize}
  \item At step $\ell$, optimize equation (\ref{eqn:probls}) with respect to $\{t_j\}$ given $\mathbf{c}^{(\ell)}$. Since $\mathbf{c}^{(\ell)}$ is known, it follows immediately that Equation~(\ref{eqn:probls}) decomposes into $M$ independent tasks, where for the $j$th task we need to find $t_j^{(\ell)}$ such that
  $C^j\widetilde{\mathcal{K}}^\T_{t_j^{(\ell)}}\mathbf{c}$ is closest
      to $\ty^j$ under the Euclidean distance. A brute force search with time
      complexity $\mathbf{\Theta}(\bbbn L)$ yields the optimal solution. If
      the time series are synchronously sampled (i.e. $C^j = \mathbb{I},
      j=1,\cdots,M$), this is equivalent to finding the shift $\tau$ corresponding the \emph{cross-correlation}, defined as
      \begin{equation}
        \label{eqn:ccd}
        \mathcal{C}(\mathbf{u}, \mathbf{v}) = \underset{\tau}{\max}\langle \mathbf{u}, \mathbf{v}_{+\tau} \rangle
      \end{equation}
      where $\mathbf{u} = \mathfrak{K}\mathbf{c}$ and $\mathbf{v} = \ty^j$ and $\mathbf{v}_{+\tau}$ refers to the vector $\mathbf{v}$ right shifted by $\tau$ positions, and where positions are shifted modulo $\bbbn$. Furthermore, as shown by~\cite{Protopapas06}, if every $\x^j$ has regular time intervals, we can use the convolution theorem to find the same value in $\mathbf{\Theta}(\bbbn\log \bbbn)$ time, that is
      \begin{equation}
        t_j^{(\ell)} = \underset{\tau}{\text{argmax}}\left(\mathscr{F}^{-1}\left[\mathscr{U}\cdot\widehat{\mathscr{V}}\right](\tau)\right)
      \end{equation}
      where $\mathscr{F}^{-1}[\cdot]$ denotes inverse Fourier transform, $\cdot$ indicates point-wise multiplication; $\mathscr{U}$ is the Fourier transform of $\mathbf{u}$ and $\widehat{\mathscr{V}}$ is the complex conjugate of the Fourier transform of $\mathbf{v}$.

  \item At step $\ell+1$, optimize equation (\ref{eqn:probls}) with respect to $\mathbf{c}^{(\ell+1)}$ given $t_1^{(\ell)},\cdots, t_M^{(\ell)}$. For the $j$th task, since $t_j^{(\ell)}$ is known,
       denote $C^j\widetilde{\mathcal{K}}^\T_{t_j^{(\ell)}}$ as $\mathfrak{M}_j^{(\ell)}$. The regularized least square problem can be reformulated as
      \begin{equation}
        \label{eqn:probrls}
        \underset{\mathbf{c}\in\bbbr^\bbbn}{\text{argmin}}\left\{
        \sum_j\gamma_{js}\|\ty^j-\mathfrak{M}_j^{(\ell)}\mathbf{c}\|^2 + \frac{1}{2} \mathbf{c}^T\mathfrak{K}\mathbf{c}\right\}.
      \end{equation}
      Taking derivatives of Equation~(\ref{eqn:probrls}), we see that the new $\mathbf{c}^{(\ell+1)}$ value is obtained by solving the following linear system
      \begin{equation}
      \label{eqn:modelsystem}
        -2\sum_j\gamma_{js}\cdot (\mathfrak{M}^{(\ell)}_j)^\T\left(\ty^j-\mathfrak{M}^{(\ell)}_j\cdot \mathbf{c}\right) + \mathfrak{K}\mathbf{c}=0.
      \end{equation}

\end{itemize}
Obviously, each step decreases the value of
the objective function and therefore the algorithm will converge.

Given the estimates of $\{\f_s\},\{t_j\}$, the optimization for $\sigma^2$ is given by
        \begin{equation}
        \label{eqn:sigma}
        \begin{split}
          \sigma^* &= \underset{\sigma\in\bbbr}{\text{argmin}}\Bigg\{\sum_jn_j\log\sigma
        + \frac{1}{2\sigma^2}\sum_j\text{Tr}(\mathbf{C}_j^g)\\
            &\qquad \frac{1}{2\sigma^2}\sum_j\sum_s\gamma_{js}
                \left(\|\y^j - [\f^*_s*\delta_{t^*_j}](\x^j) - \boldsymbol\mu_{js}\|^2\right)\Bigg\}
        \end{split}
        \end{equation}
        where $\{\f^*\}$ and $\{t^*_j\}$ are obtained from the previous optimization steps. Let $R=\sum_j\text{Tr}(\mathbf{C}_j^g)+\sum_j\sum_s\gamma_{js}
    \left(\|\y^j - [\f^*_s*\delta_{t_j}](\x^j) - \boldsymbol\mu_{js}\|^2\right)$. Then it is easy to see that $(\sigma^*)^2 = R/\sum_jn_j$.

    \subsubsection{Learning the kernel for individual effect}
    \cite{lu2008rkh} have already shown how to optimize the kernel function in a similar context. Here we provide some of the details for completeness. If the kernel function $\mathcal{K}$ admits a parametric form with parameter $\theta$, for example the RBF kernel
    \begin{equation}
    \label{eqn:rbf}
      \mathcal{K}(x,y) = a\exp\left\{-\frac{\|x-y\|^2}{2s^2}\right\}
    \end{equation}
    where $\theta=\{a,s\}$, then the optimization of the kernel $\mathcal{K}$ amounts to finding $\theta^*$ such that
    \begin{equation}
        \label{eqn:opk}
        \begin{split}
      \theta^* = \underset{\theta}{\text{argmax}}\Bigg\{-\frac{1}{2}\sum_j \log|(\mathbf{K}_j;\theta)|
      - \frac{1}{2}\sum_j\sum_s\gamma_{js}
        \text{Tr}\left(
        (\mathbf{K}_j;\theta)^{-1}(\mathbf{C}_{j}^g + \boldsymbol\mu_{js}^g(\boldsymbol\mu_{js}^g)^\T)\right)\Bigg\}.
        \end{split}
    \end{equation}
    It is easy to see the gradient of the right hand side of Equation~(\ref{eqn:opk}) is
    \begin{equation}
        \begin{split}
      -\frac{1}{2}\sum_j \text{Tr}\left(\mathbf{K}_j\frac{\partial\mathbf{K}_j}{\partial{\theta}}\right)- \frac{1}{2}\sum_j\sum_s\gamma_{js}
        \text{Tr}\left(
        \mathbf{K}_j^{-1}\frac{\partial\mathbf{K}_j}{\partial{\theta}}\mathbf{K}_j^{-1}(\mathbf{C}_{j}^g + \boldsymbol\mu_{js}^g(\boldsymbol\mu_{js}^g)^\T)\right).
        \end{split}
    \end{equation}
Therefore, any optimization method, e.g.\ conjugated gradients can
be utilized to find the optimal parameters. Notice that given the
inverse of kernel matrix $\{\mathbf{K}_j\}$, the computation of the
derivative requires $\Theta(\sum n_j^2)$ steps. The parametric form of the
kernel is a prerequisite to perform the regression task when
examples are not sampled synchronously as in our development above.

    If the data is synchronously sampled, for classification tasks we only need to find the kernel matrix $\mathbf{K}$ for the given sample points and the optimization problem can be rewritten as
    \begin{equation}
    \begin{split}
      \mathbf{K}^* &= \underset{\mathbf{K}}{\text{argmax}}\Bigg\{-\frac{1}{2}\sum_j \log|\mathbf{K}| - \frac{1}{2}\sum_j\sum_s\gamma_{js}
        \text{Tr}\left(
        \mathbf{K}^{-1}(\mathbf{C}_{j}^g + \boldsymbol\mu_{js}^g(\boldsymbol\mu_{js}^g)^\T)\right)\Bigg\}.
    \end{split}
    \end{equation}
    Similar to maximum likelihood estimation for multivariate Gaussian distribution, the solution is
    \begin{equation}
        \label{eqn:nonp}
        \mathbf{K}^* = \frac{1}{M}\sum_j\sum_s\gamma_{js}(\mathbf{C}_{j}^g + \boldsymbol\mu_{js}^g(\boldsymbol\mu_{js}^g)^\T).
    \end{equation}

In our experiments, we
use both approaches where for the parametric form we use
the RBF kernel as outlined above.

\subsection{Algorithm Summary}

The various steps in our algorithm and their time complexity are summarized in
Algorithm~\ref{alg:gmt}.

\algsetup{indent=2em} 
\begin{algorithm}[h!] \caption{\sc EM algorithm for Shift-invariant GMT}\label{alg:gmt}
\begin{algorithmic}[1]
\STATE Initialize $\{f_s^{(0)}\}, \{t_j^{(0)}\}, \alpha^{(0)}$ and
$\mathcal{K}^{(0)}$. \REPEAT \STATE Calculate $\fK_j^{(t)}$
according to $\x^j, \mathcal{K}^{(t-1)}$. The time complexity for
constructing kernel are $\Theta(\sum n_j^2)$ and $\Theta(1)$ in
parametric and nonparametric case respectively. \STATE Calculate
$\gamma_{js}$ according to Equation~(\ref{eqn:gammajs}). For each task,
we need to invert the covariance matrix in the marginal distribution
and then calculate the likelihood, thus the time complexity is
$\Theta(\sum n_j^3)$. \FORALL{$s$ such that $0\leq s\leq k$} \STATE
Update $\alpha^{(t)}$ such that $\alpha^{(t)}_s =
\sum_j\gamma_{js}/M$. \REPEAT \STATE Update $\{t_j\}$ w.r.t. cluster
$s$ such that $t_j\in\{\tilde{t}_1,\cdots,\tilde{t}_L\}$ and
minimize
$\|\ty^j-C^j\cdot\widetilde{\mathcal{K}}^\T_{t_j}\mathbf{c}_s^{(0)}\|^2$.
The time complexity is $\Theta(L\bbbn)$ as discussed above. \STATE
Update $\mathbf{c}_s^{(t+1)}$ by solving linear system Equation~(\ref{eqn:modelsystem}), which requires $\Theta(\bbbn^3)$.
\UNTIL{converges \OR reach the iteration limit} \ENDFOR \STATE
Update $\sigma^{(t+1)}$ according to Equation~(\ref{eqn:sigma}). \STATE
Update the parameters of the kernel or the kernel matrix directly
via optimizing Equation~(\ref{eqn:opk}) or using the closed-form solution
Equation~(\ref{eqn:nonp}) for $\mathbf{K}$. In the former case, a
gradient based optimizer can be used with time complexity
$\Theta(\sum n_j^2)$ for each iteration; while in the later case,
the estimation only requires $\Theta(kM\bbbn)$.

\UNTIL{converges or reach the iteration limit}
\end{algorithmic}
\end{algorithm}

Once the model parameters $\mathcal{M}$ are learned (or if they are given in
advance), we can use the model to perform regression or classification
tasks. The following summarizes the procedures used in our experiments.

\begin{itemize}
  \item \textbf{Regression: }
To predict a new sample point for an existing task (task $j$) we
calculate its most likely cluster assignment $z_j$ and then predict the $y$
value based on this cluster. Concretely,
$z_j$ is determined by
  \begin{equation}
    z_j = \underset{s=\{1,\cdots,k\}}{\text{argmax}}\left[\Pr(z_j =
    s|\x^j, \y^j; \mathcal{M})\right]
  \end{equation}
and
given a new data point $x$,
the prediction $y$ is given by
  \begin{displaymath}
    y = \left[\f_{z_j}*\delta_{t_j}\right](x) + \tilde{f}^j(x).
  \end{displaymath}
  \item \textbf{Classification: }
For classification, we get a new time series and want to predict its label.
Recall from Section 3, Equation~(\ref{eqn:class}) that we learn a separate model
for each class and  predict using
\[
      o =
      \underset{\ell=\{1,\cdots,L\}}{\text{argmax}}\Pr(\y|\x; \mathrm{M}_\ell)\Pr(\ell).
\]
In this context,
$\Pr(\ell)$ is
  estimated by the frequencies of each class and the likelihood portion is
  given by first finding the best time shift $t$ for the new time series and then calculating the likelihood according to
  \begin{equation}
  \begin{split}
    \Pr(\y|\x; M_{\ell}) =
    \sum_{z}\Pr(z|\mathcal{M}_{\ell})\Pr(\y|z,
    \x;\mathcal{M}_{\ell})
  \end{split}
  \end{equation}
  where $\mathcal{M}_\ell$ is the learned parameter set and the second term is calculated via Equation~(\ref{eqn:14p}).
\end{itemize}

\section{Infinite mixture of Gaussian processes}
\label{sec:infinite}
In this section we develop an extension of the model removing the assumption
that the number of centers $k$ is known in advance.

\subsection{Dirichlet process basics}
We start by reviewing basic concepts for Dirichlet processes.
Suppose we have i.i.d.\ data such that
\begin{displaymath}
x_1, x_2, \cdots, x_n\sim \mathcal{F}
\end{displaymath}
where $\mathcal{F}$ is an unknown distribution that needs to be
inferred from $\{x_i\}$. A Parametric Bayesian approach assumes
$\mathcal{F}$ is given by a parametric family $\mathcal{F}_\theta$
and the parameters $\theta$ follow a certain distribution that comes from our
prior belief. However, this assumption has limitations both in the scope and the type
of inferences that can be performed. Instead, nonparametric Bayesian
approach places a prior distribution on the distribution $\mathcal{F}$ directly.
The Dirichlet process (DP) is used for such purpose. The DP is
parameterized by a base distribution $G_0$ and a positive scaling
parameter (or concentration parameter) $\alpha$. A random measure $G$ is distributed according to
a DP with base measure $G_0$ and scaling parameter $\alpha$ if for
all finite measurable partitions $\{B_i\}, i=1,\ldots,k$,
\begin{displaymath}
  (G(B_1), G(B_2), \cdots, G(B_k)) \sim \text{Dir}(\alpha G_0(B_1), \alpha
  G_0(B_2), \cdots, \alpha G_0(B_k))
\end{displaymath}
where $\text{Dir}(\cdot)$ is the Dirichlet distribution.
It is known that $G$ is almost surely a discrete measure.

The Dirichlet process mixture model extends this setting, where
the DP is used as a nonparametric prior in a hierarchical Bayesian
specification. More precisely,
\begin{displaymath}
  \begin{split}
    G|\{\alpha, G_0\} &\sim \mathcal{DP}(\alpha,G_0)\\
    \eta_n|G &\sim G \ \ \ \ \ \ \ \ n=1,2,\ldots\\
    x_n|\eta_n &\sim f(x_n|\eta_n)
  \end{split}
\end{displaymath}
where $f$ is some probability density function that is parameterized
by $\eta$. Data generated from this model can be naturally partitioned
according to the distinct values of the parameter $\eta_n$. Hence, the DP
mixture can be interpreted as a mixture model where the number of mixtures is
flexible and grows as the new data is observed. Alternatively, we
can view the infinite mixture model as the limit of the finite
mixture model. Consider the Bayesian finite mixture model with a
symmetric Dirichlet distribution as the prior of the mixture
proportions. When the number of mixtures $k\rightarrow\infty$,
the Dirichlet distribution becomes a Dirichlet
process~\citep[see][]{neal2000markov}.

\cite{sethuraman1994constructive} provides a more explicit
construction of the DP which is called the \emph{stick-breaking
construction} (SBC). Given $\{\alpha, G_0\}$, we have two
collections of random variables $V_i\sim\text{Beta}(1,\alpha)$ and
$\eta_i^*\sim G_0,\quad i = \{1,2,\cdots,\}$. The SBC of $G$ is
\begin{displaymath}
  \begin{split}
    \pi_i(\mathbf{v}) &= v_i\prod_{j=1}^{i-1}(1-v_j)\\
    G &= \sum_{i=1}^\infty\pi_i(\mathbf{v})\delta_{\eta_i^*}.
  \end{split}
\end{displaymath}
If we set $v_K = 1$ for some $K$, then we get a truncated
approximation to the DP
\begin{displaymath}
G = \sum_{i=1}^K\pi_i(\mathbf{v})\delta_{\eta_i^*}.
\end{displaymath}
\cite{ishwaran2001gibbs} shows that when selecting the truncation
level $K$ appropriately, the truncated DP behaves very similarly to the
original DP.

\subsection{The DP-GMT Model and Inference Algorithm}
\begin{figure}
\centering
        \includegraphics[width=5cm]{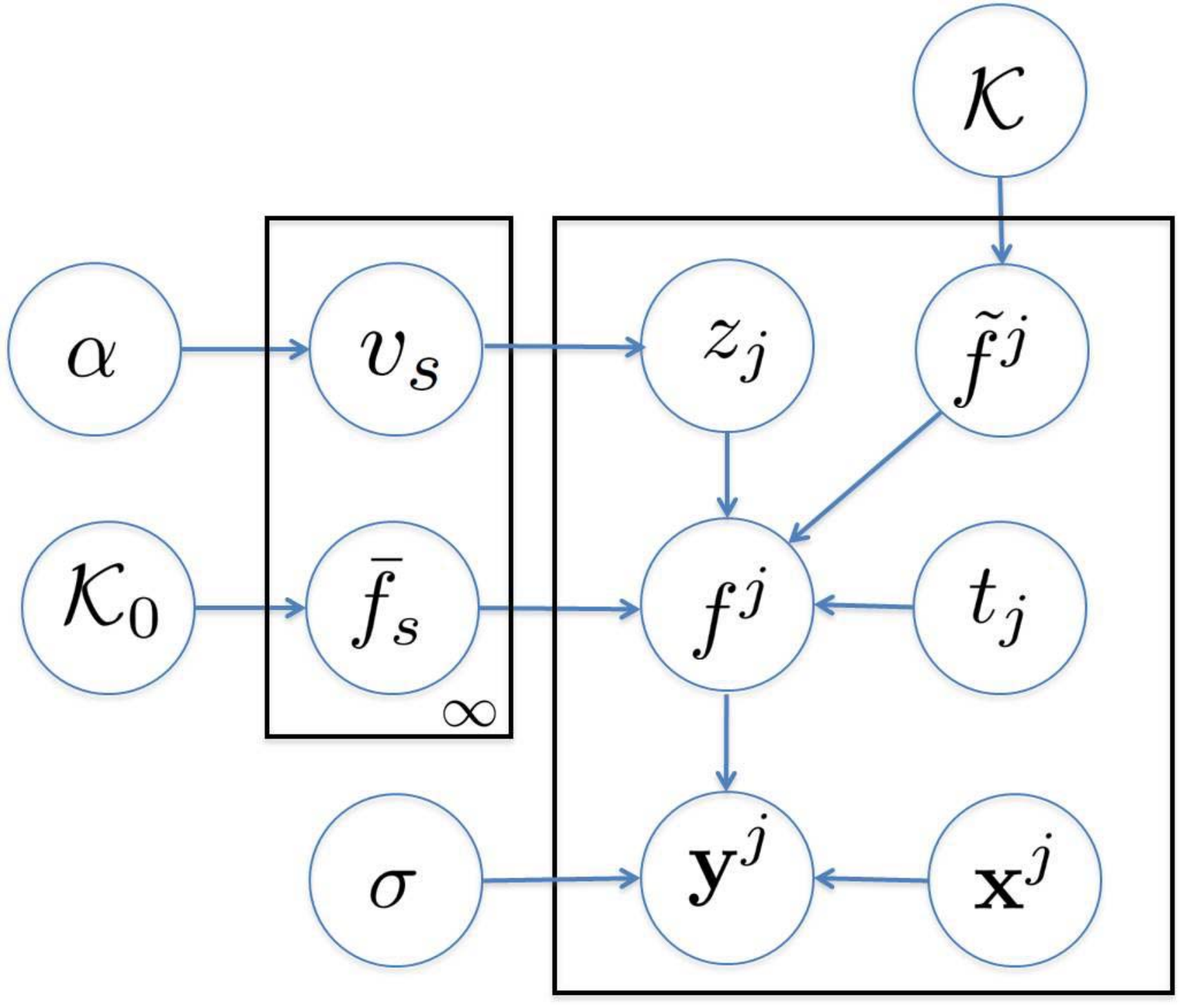}
        \caption{DPGMT: Plate graph}
        \label{fig:dpgmt_graph}
\end{figure}

In this section, we extend our model by modeling the mixture proportions
using a DP prior. The plate graph is shown in
Figure~\ref{fig:dpgmt_graph}. Under the SBC,
the generative process is as follows
\begin{enumerate}
\item Draw $v_s|\alpha\sim\text{Beta}(1,\alpha)$, $s = \{1,2,\ldots\}$
\item Draw $\f_s|\mathcal{K}_0\sim \exp\left\{-\frac{1}{2}\|\f_s\|_{\mathcal{H}_0}^2\right\}$, $s = \{1,2,\ldots\}$
\item For the $j$th time series
\begin{enumerate}
    \item Draw $z_j|\{v_1,v_2,\ldots\}\sim \text{Discrete}(\pi(\mathbf{v}))$, where $\pi_s(\mathbf{v}) = v_s\prod_{i = 1}^{s-1}(1-v_i)$;
    \item Draw $\tilde{f}^j|\mathcal{K} \sim \exp\left\{-\frac{1}{2}\|\tilde{f}^j\|_\mathcal{H}^2\right\}$;
    \item Draw $\y^j|z_j, f^j, \x^j, t_j, \sigma^2 \sim \mathcal{N}\left(f^j(\x^j), \sigma^2\mathbb{I}\right)$, where $f^j = \f_{z_{j}}*\delta_{t_j}+\tilde{f}^j$.
\end{enumerate}
\end{enumerate}
In this model, the concentration parameter $\alpha$ is assumed to be
known. As in Equation~(\ref{eqn:EM}), the inference task is to
find the MAP estimates of the parameter set $\mathcal{M}=\{\{\f_s\},
\{t_j\}, \sigma^2, \mathcal{K}\}$. Notice that in contrast with the
previous model, the mixture proportion are not estimated here.
To perform the inference, we must consider another set of
hidden variables $\mathbf{v}=\{v_i\}$ in addition to $\mathbf{f}$ and
$\mathbf{z}$. However, calculating the posterior of the hidden
variables is intractable, thus the variational \textsc{em} algorithm
\citep[e.g.,][]{bishop2006pattern}
is used
to perform the approximate inference. The algorithm can be
summarized as follows:
\begin{itemize}
\item \textbf{Variational E-Step} Choose a family $\mathcal{G}$ of variational distributions $q(\mathbf{f,v,z})$
and find the distribution $q^*$ that minimizes the Kullback-Leibler
(\textbf{KL}) divergence between the posterior distribution and the
proposed distribution given the current estimate of parameters, i.e.
\begin{equation}
\label{eqn:kl} q^*(\mathbf{f,v,z}; \mathcal{M}^g) =
\underset{q\in\mathcal{G}}{\text{argmin}}\
\text{\textbf{KL}}\left(q(\mathbf{f,v,z})||\Pr(\mathbf{f,v,z}|\mathcal{X,Y};\mathcal{M}^g)\right)
\end{equation}
where
\begin{displaymath}
\text{\textbf{KL}}\left(q(\mathbf{f,v,z})||\Pr(\mathbf{f,v,z}|\mathcal{X,Y};\mathcal{M}^g)\right) = \int
\log\left[\frac{q(\mathbf{f,v,z})}{\Pr(\mathbf{f,v,z}|\mathcal{X,Y};\mathcal{M}^g)}\right]\text{d}q(\mathbf{f,v,z}).
\end{displaymath}
\item \textbf{Variational M-Step}
Optimize the parameter set $\mathcal{M}$ such that
\begin{displaymath}
\mathcal{M}^* = \underset{\mathcal{M}}{\text{argmax}}\ Q(\mathcal{M, M}^g)
\end{displaymath}
where
\begin{equation}
\label{eqn:vbq}
Q(\mathcal{M},\mathcal{M}^g) = \bbbe_{q^*(\mathbf{z,f,v};
    \mathcal{M}^g)}\left[\log\left\{\Pr(\mathcal{Y},\mathbf{f,z,v}|\mathcal{X}; \mathcal{M})\times \Pr[\{\f_s\};\mathcal{K}_0]\right\}\right].
\end{equation}
\end{itemize}

\textbf{Variational E-Step}.
For the variational distribution $q()$ we use the
\emph{mean field approximation}~\citep{wainwright2008graphical}.
That is we assume a
factorized distribution for disjoint group of random variables.
This results in an analytic tractable optimization problem.
In addition, following ~\citep{blei2006variational},
we approximate the distribution over $\mathbf{v}$
using
a truncated
stick-breaking representations, where for a fix $T$, $q(v_T = 1) =
1$ and therefore $\pi_s(\mathbf{v}) = 0,\ s > T$. In this paper, we
fix the truncation level $T$ while in general it can also be treated as a
variational parameter. Concretely, we propose the following factorized
family of variational distributions over the hidden variables
$\{\mathbf{f,v,z}\}$:

\begin{equation}
q(\mathbf{f,v,z}) = \prod_{s=1}^{T-1}q_s(v_s)\prod_{j=1}^{M}q_j(f_j,
z_j).
\end{equation}
Note that we do not assume any parametric form for $\{q_s, q_j\}$
and our only assumption is that the distribution factorizes into
independent components. To optimize Equation~(\ref{eqn:kl}), recall the
following result from~\cite[][Chapter 8]{bishop2006pattern}:
\begin{lemma}
Suppose we are given a probabilistic model with a joint distribution
$\Pr(\mathbf{X,Z})$ over $\mathbf{X,Z}$ where
$\mathbf{X}=\{\mathbf{x}_1, \mathbf{x}_2, \cdots, \mathbf{x}_N\}$
denote the observed variables and all the parameters and hidden
variables are $\mathbf{Z}=\{\mathbf{z}_1, \mathbf{z}_2, \cdots,
\mathbf{z}_M\}$. Assume the distribution of $\mathbf{Z}$ has the
following form:
\begin{displaymath}
q(\mathbf{Z}) = \prod_i^M q_i(z_i).
\end{displaymath}
Then, the \textbf{KL} divergence between the posterior distribution $\Pr(\mathbf{Z}|\mathbf{X})$ and $q(\mathbf{Z})$ is minimized and the optimal solution $q_j^*(\mathbf{z}_j)$ is given by
\begin{displaymath}
q_j^*(\mathbf{z}_j) \propto \text{exp}\left(\bbbe_{i\neq
j}\left[\log\Pr(\mathbf{X}, \mathbf{Z})\right]\right)
\end{displaymath}
where $\bbbe_{i\neq j}[\cdots]$ denotes the expectation w.r.t. $q()$
over all $Z_j,\ j\neq i$.
\end{lemma}
From the graphical model in Figure~\ref{fig:dpgmt_graph}, the joint
distribution of $\Pr(\mathcal{Y},
\mathbf{f,v,z}|\mathcal{X};\mathcal{M}^g)$ can be written as:
\begin{displaymath}
\begin{split}
\Pr(\mathcal{Y}, \mathbf{f, v, z} | \mathcal{X}) & = \Pr(\mathcal{Y}|\mathcal{X}, \mathbf{f,z})\Pr(\mathbf{z}|\mathbf{v})\Pr(\mathbf{f}|\mathcal{X})\Pr(\mathbf{v}|\alpha)\\
&= \prod_j\Pr(\y^j|\x^j, \mathbf{f}_j, z_j)\prod_j\Pr(z_j|\mathbf{v})\prod_j\Pr(\mathbf{f}_j|\x^j)\prod_s\Pr(v_s|\alpha).
\end{split}
\end{displaymath}
Equivalently,
\begin{displaymath}
\begin{split}
\log\Pr(\mathcal{Y}, \mathbf{f, v, z} | \mathcal{X}) &=
\sum_j\log\Pr(\y^j|\x^j, \mathbf{f}_j, z_j) \\
&\qquad+ \sum_j\log\Pr(z_j|\mathbf{v}) +
\sum_j\log\Pr(\mathbf{f}_j|\x^j) + \sum_s\log\Pr(v_s|\alpha).
\end{split}
\end{displaymath}
First we consider the distribution of $q_s(\mathbf{v})$.
Following~\cite{blei2006variational}, the second term
can be expanded as
\begin{equation}
\label{eqn:logzj} \log\Pr(z_j|\mathbf{v}) = \sum_{t =
1}^T\mathbbold{1}_{\{z_j
> t\}}\log(1 - v_t) + \mathbbold{1}_{\{z_j = t\}}\log v_t
\end{equation}
where $\mathbbold{1}$ is the indicator function. Therefore, using the lemma
above and denoting
$\mathbf{v}\backslash v_s$ by $\mathbf{v}_{-s}$, we have
\begin{displaymath}
\begin{split}
\log q_s(v_s) &\propto \bbbe_{\mathbf{z,f,v}_{-s}}\left[\log\Pr(\mathcal{Y}, \mathbf{f, v, z} | \mathcal{X})\right]\\
& =
\sum_j\left(\bbbe_{\mathbf{z,f,v}_{-s}}[\mathbbold{1}_{\{z_j
> s\}}]\log(1 - v_s) + \bbbe_{\mathbf{z,f,v}_{-s}}[\mathbbold{1}_{\{z_j
= s\}}]\log
v_s\right) + \log\Pr(v_s|\alpha) + \text{constant}\\
&= \sum_j\left(q(z_j > s)\log(1 - v_s) + q(z_j = s)\log
v_s\right) + \log\Pr(v_s|\alpha) + \text{constant}
\end{split}
\end{displaymath}

Recalling that the prior is given by $\text{Beta}(1,\alpha)$ we see that
the distribution of $q_s(v_s)$ is
\begin{displaymath}
q_s(v_s) \propto v_s^{\sum_jq(z_j = s)}(1-v_s)^{\alpha +
\sum_j\sum_{l=s+1}^Tq(z_j = l) - 1}.
\end{displaymath}
Observing the form of $q_s(v_s)$, we can see that it is a Beta
distribution and $q_t(v_t)\sim\text{Beta}(\gamma_{t,1},
\gamma_{t,2})$ where
\begin{displaymath}
\begin{split}
\gamma_{t,1} &= 1 + \sum_jq(z_j = t)\\
\gamma_{t,2} &= \alpha + \sum_j\sum_{l=s+1}^Tq(z_j = l).
\end{split}
\end{displaymath}

We next consider $q_j(\mathbf{f}_j, z_j)$. Notice that we can always
write $q_j(\mathbf{f}_j, z_j) = q_j(\mathbf{f}_j|z_j)q_j(z_j)$.
Denote $h(z_j) =
\bbbe_{\mathbf{v}}\left[\log\Pr(z_j|\mathbf{v})\right]$, then
again using the lemma above we have
\begin{displaymath}
\begin{split}
q_j(\mathbf{f}_j|z_j)q_j(z_j) &\propto e^{h(z_j)}\Pr(\y^j|\mathbf{x}^j,\mathbf{f}_j, z_j)\Pr(\mathbf{f}_j|\x^j)\\
&= e^{h(z_j)}\Pr(\y^j, \mathbf{f}_j|\x^j, z_j)\\
&\propto e^{h(z_j)}\Pr(\y^j|\x^j, z_j)\Pr(\mathbf{f}_j|\x^j, \y^j, z_j)\\
 &\propto \left[\underbrace{e^{h(z_j)}\Pr(\y^j|\x^j, z_j)}_{q_j(z_j)}\right]\left[\underbrace{\Pr(\mathbf{f}_j|\x^j, \y^j, z_j)}_{q_j(\mathbf{f}_j|z_j)}\right].\\
\end{split}
\end{displaymath}
The equality in the second line holds because
$\Pr(\mathbf{f}_j|\x^j) = \Pr(\mathbf{f}_j|\x^j, z_j)$; their
distributions become coupled when conditioned on the observations
$\y^j$, but without such observations they are independent.
Therefore the left term yields

\begin{displaymath}
q_j(z_j)  \propto e^{h(z_j)}\Pr(\y^j |\x^j, z_j)
\end{displaymath}
where $\Pr(\y^j|\x^j, z_j)$ is given by Equation~(\ref{eqn:14p}).
The value of $h(z_{j})$ can be calculated using
Equation~(\ref{eqn:logzj}):
\begin{displaymath}
\begin{split}
\log\Pr(z_j = s|\mathbf{v}) & = \sum_{t =1}^{s-1} \log(1 - v_t) + \log v_s \\
    h(z_j = s)
&= \bbbe_{v_s}[\log v_s] + \sum_{i = 1}^{s-1}\bbbe_{v_i}\left[\log(1-
v_i)\right]
\end{split}
\end{displaymath}
where
\begin{displaymath}
\label{eqn:gammai12}
\begin{split}
\bbbe_{v_t}[\log v_t] &= \Psi(\gamma_{i,1}) - \Psi(\gamma_{i,1} + \gamma_{i,2})\\
\bbbe_{v_i}\left[\log(1- v_i)\right] &= \Psi(\gamma_{i,2}) -
\Psi(\gamma_{i,1} + \gamma_{i,2}).
\end{split}
\end{displaymath}
 Consequently, $q_j(z_j)$
has the following form
\begin{equation}
\label{eqn:vz} q_j(z_j = t) \propto
\text{exp}\left\{\bbbe_{v_t}[\log v_t] + \sum_{i =
1}^{t-1}\bbbe_{v_i}\left[(1- v_i)\right]\right\}\times
\mathcal{N}\left(\overline{\mathbf{f}}^j, \mathbf{K}_j^g +
\sigma^2\mathbb{I}\right).
\end{equation}
Note that this is the same form as in Equation~(\ref{eqn:mum}) of the previous model
where $\Pr(z_j;\mathcal{M}^g)$ is replaced by $e^{h(z_{j} = t)}$.

Given $z_j$,  $q_j(\mathbf{f}_j|z_j)$  is identical to Equation~(\ref{eqn:fjconditional}) and leads to the
conditional distribution such that
\begin{displaymath}
  q_j(\mathbf{f}_j|z_j) \propto \Pr( \y^j|\x^j,\mathbf{f}_j, z_j)\Pr(\mathbf{f}_j;\x^j)
\end{displaymath}
which is the posterior distribution under GP regression and thus is
exactly the same form as in the previous model.

\textbf{Variational M-Step}. Denote $\widetilde{Q}$ as
the expectation of the complete data log likelihood w.r.t. the hidden
variables. Then as in Equation~(\ref{eqn:tQ}), we have
\begin{equation}
      \label{eqn:tQv}
      \begin{split}
        \widetilde{Q} &=\bbbe_{q(\mathbf{v})}\bbbe_{q(\mathbf{z})}\bbbe_{q(\mathbf{f}|\mathbf{z})}
        \log\left(\prod_j\prod_s\left[\pi_s(\mathbf{v})\Pr(\y^j|\mathbf{f}_j,z_j=s;\mathcal{M})
        \Pr(\mathbf{f}_j;\mathcal{M})\right]^{z_{js}}\right)\\
        &= \bbbe_{\mathbf{v}}\Bigg[\sum_{\mathbf{z}}q(\mathbf{z})\Bigg\{\sum_j\sum_sz_{js}
        \cdot\int dq(\mathbf{f}_j|z_j=s)\\
        &\qquad \times\log\left[\pi_s(\mathbf{v})\Pr(\y^j|\mathbf{f}_j,z_j=s;\mathcal{M})
        \Pr(\mathbf{f}_j;\mathcal{M})\right]\Bigg\}\Bigg]\\
        &= \sum_{\mathbf{z}}q(\mathbf{z})\Bigg\{\sum_j\sum_sz_{js}\cdot\int d
        q(\mathbf{f}_j|z_j=s)\\
        &\qquad \times \log\left[\Pr(\y^j|\mathbf{f}_j,z_j=s;\mathcal{M})
        \Pr(\mathbf{f}_j;\mathcal{M})\right]\Bigg\} + \bbbe_{\mathbf{v}}\left[\sum_j\sum_s\log\pi_s(\mathbf{v})\right].
      \end{split}
    \end{equation}
Notice that
$\bbbe_{\mathbf{v}}\left[\sum_j\sum_s\log\pi_s(\mathbf{v})\right]$
is a constant w.r.t.\  the parameters of $\mathcal{M}$
and can be dropped in the optimization.
Thus, following the same derivation as in the GMT model,
we have the form of the $Q$
function as
  \begin{equation}
  \begin{split}
    Q(\mathcal{M},\mathcal{M}^g) &=
    -\frac{1}{2}\sum_s\|\f_s\|_{\mathcal{H}_0}^2 - \sum_jn_j\log\sigma \\
    &\qquad - \frac{1}{2\sigma^2}\sum_j\sum_s\gamma_{js} \bbbe_{\{q(\mathbf{f}_j|z_j=s)\}}\left[\|\y^j - [\f_s*\delta_{t_j}](\x^j) - \mathbf{f}_j\|^2\right]\\
    &\qquad + \sum_j\sum_s\gamma_{js}\bbbe_{\{q(\mathbf{f}_j)\}}\left[\log \Pr(\mathbf{f}_j;\mathcal{M})\right].
  \end{split}
  \end{equation}
where $\gamma_{js}$ is given by Equation~(\ref{eqn:gammajs}). Now because the
$q_j(z_j)$ and $q_j(f_j|z_j)$ have exactly the same form as before (except
$\Pr(z_j;\mathcal{M}^g)$ is replaced by Equation~(\ref{eqn:vz})),
the previous derivation of the \textbf{M-Step} w.r.t. the parameter
set $\mathcal{M}$ still holds.

To summarize, the algorithm is the same as Algorithm 1 except that
\begin{itemize}
\item
we drop
step 6,
\item
we add a step between steps 3 and 4 calculating $\gamma_{i,1}$
and $\gamma_{i,2}$ using Equation~\ref{eqn:gammai12},
\item
step 4 calculating
Equation~(\ref{eqn:gammajs}) uses Equation~(\ref{eqn:vz}) instead of
Equation~(\ref{eqn:mum}).
\end{itemize}

\section{Experiments}
Our implementation of the algorithm makes use of the gpml
package~\citep{rasmussen2010gaussian} and extends it to implement the required
functions. The \textsc{em} algorithm is restarted 5 times and the
function that best fits the data is chosen. The \textsc{em}
algorithm stops when difference of the log-likelihood is less than
10e-5 or at a maximum of 200 iterations.
\subsection{Regression on Synthetic data}
In the first experiment, we demonstrate the performance of our algorithm on a regression task with artificial data. We generated the data following Assumption
1 under a mixture of three Gaussian processes. More precisely, each
$\f_s(x),s=1,2,3$ is generated on the interval $[-50, 50]$ from a Gaussian
process with covariance function
\begin{displaymath}
  \text{cov}[\f_s(t_1), \f_s(t_2)] = e^{-\frac{(t_1-t_2)^2}{25}},\quad s=1,2,3.
\end{displaymath}
The individual effect $\tilde{f}_j$ is sampled via a Gaussian process
with the covariance function
\begin{displaymath}
  \text{cov}[\tilde{f}_j(t_1), \tilde{f}_j(t_2)] = 0.2e^{-\frac{(t_1-t_2)^2}{16}}.
\end{displaymath}
Then the hidden label $z_j$ is sampled from a discrete distribution with the parameter $\alpha=[0.5,0.5]$. The vector $\breve{\x}$ consists of 100 samples on $[-50, 50]$\footnote{The samples are generated via Matlab command: linspace(-50,50,100).}. We fix a sample size $N$, each $\x^j$ includes $N$ randomly chosen points from
$\{\breve{x}_1,\cdots, \breve{x}_{100}\}$ and the
observation $f^j(\x^j)$ is obtained as
$(f_{z_j}+\tilde{f}_j)(\x^j)$.
In the experiment, we vary the individual sample
length $N$ from 5 to 50. Finally, we generated 50 random tasks with
the observation $\y^j$ for task $j$ given by
\begin{displaymath}
  \y^j \sim \mathcal{N}(f^j(\x^j), 0.01\times\mathbb{I}),\quad j=1,\cdots, 50.
\end{displaymath}
The methods compared here include
\begin{enumerate}
  \item \textbf{Single-task learning procedure (ST)}, where each $\f^j$ is estimated only using $\{\x^j_i, \y^j_i\}, i=1,2,\cdots,N$.
  \item \textbf{Single center mixed-effect multi-task learning (SCMT)}, amounts to the mixed-effect model~\citep{pillonetto2010bayesian} where one average function $\f$ is learned
  from $\{\x^j,\y^j\}, j=1,\cdots,50$ and $f^j = \f+\tilde{f}^j, j=1,\cdots,50$.
  \item \textbf{Grouped mixed-effect model (GMT)}, the proposed method with number of clusters fixed to be the true model order.
  \item \textbf{Dirichlet process Grouped mixed-effect model
  (DP-GMT)}, the infinite mixture extension of the proposed model.
  \item \textbf{``Cheating'' grouped fixed-effect model (CGMT)}, which follows the same algorithm as the grouped mixed-effect model but uses the true label $z_j$ instead of their expectation for each task $j$. This serves as an upper bound for the performance of the proposed algorithm.
\end{enumerate}
All algorithms (except for \textbf{ST} which does not estimate the kernel of the individual variations)
use the same method to learn the kernel of the individual effects, which is
assumed to have the form
\begin{displaymath}
  \text{cov}[\tilde{f}_j(t_1), \tilde{f}_j(t_2)] = ae^{-\frac{(t_1-t_2)^2}{s^2}}.
\end{displaymath}
\begin{figure}
\centering
        \includegraphics[width=10cm]{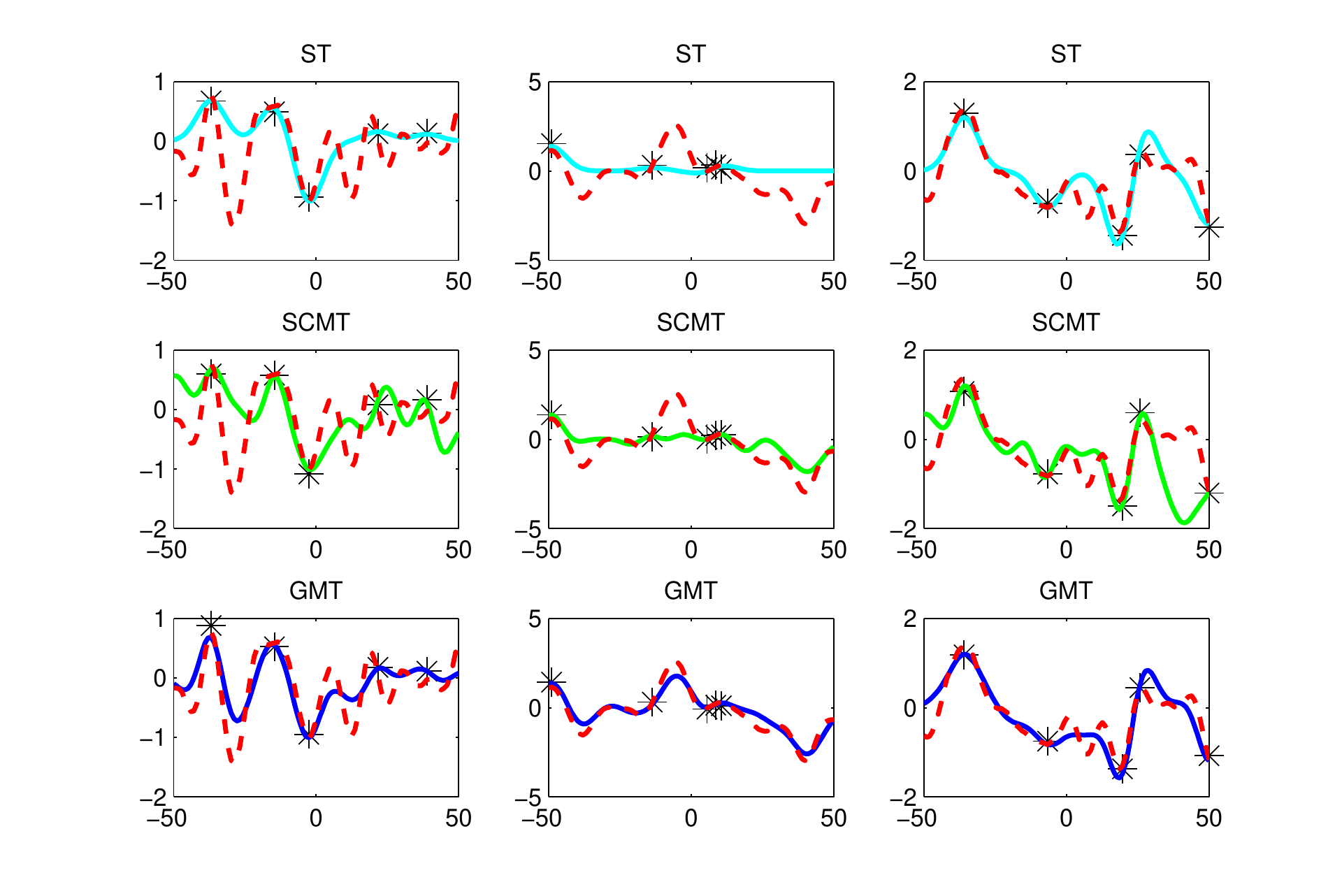}
        \caption{Simulated data: Comparison of the estimated function between single, multi-task and  grouped multi-task. The red dotted line is the reference true function.}
        \label{fig:examples}
\end{figure}
The Root Mean Square Error (RMSE) for the four approaches is
reported. For task $j$, the RMSE is defined as
\begin{displaymath}
  \text{RMSE}_j=\sqrt{\frac{1}{100}\|f(\breve{\x}) - f^j(\breve{\x})\|^2}
\end{displaymath}
where $f$ is the learned function and RMSE for the data set is the mean of
$\{\text{RMSE}_j\},j=1,\cdots,50$.
To illustrate the results qualitatively, we first plot in
Figure~\ref{fig:examples} the true and learned functions in one trial.
The left/center/right column illustrates some task that is sampled from group
effect $\f_1$, $\f_2$ and $\f_3$. It it easy to see that, as expected, the tasks are
poorly estimated under ST due the sparse sampling.
The SCMT performs better than ST but its estimate is poor in areas where the three centers disagree. The estimates of GMT are much closer to the true function.

\begin{figure}
        \centering
        \includegraphics[width=8cm]{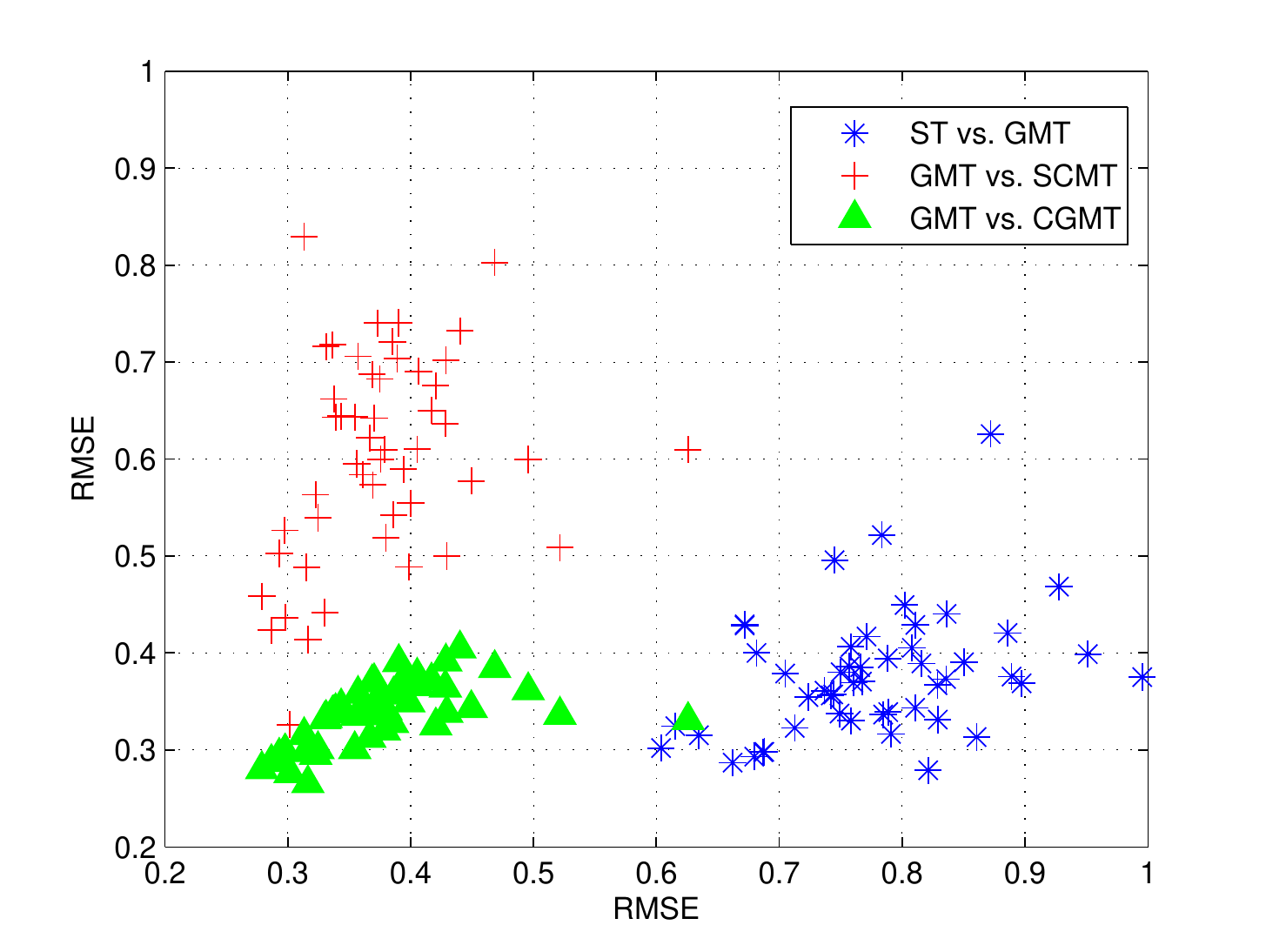}
        \caption{Simulated data: Comparison between single, multi-task, and grouped multi-task
        when sample size is 5. The figure gives 3 pairwise comparison. The Blue stars denote ST vs. GMT: we can see the GMT is better than ST since the stars are concentrated on the
        lower right. Similarly, the plot of red pluses demonstrates the advantage of GMT over SCMT and
        the plot of green triangles shows that the algorithm behaves almost as well as its upper bound.}
        \label{fig:1a}
\end{figure}

\begin{figure}
        \centering
        \includegraphics[width=8cm]{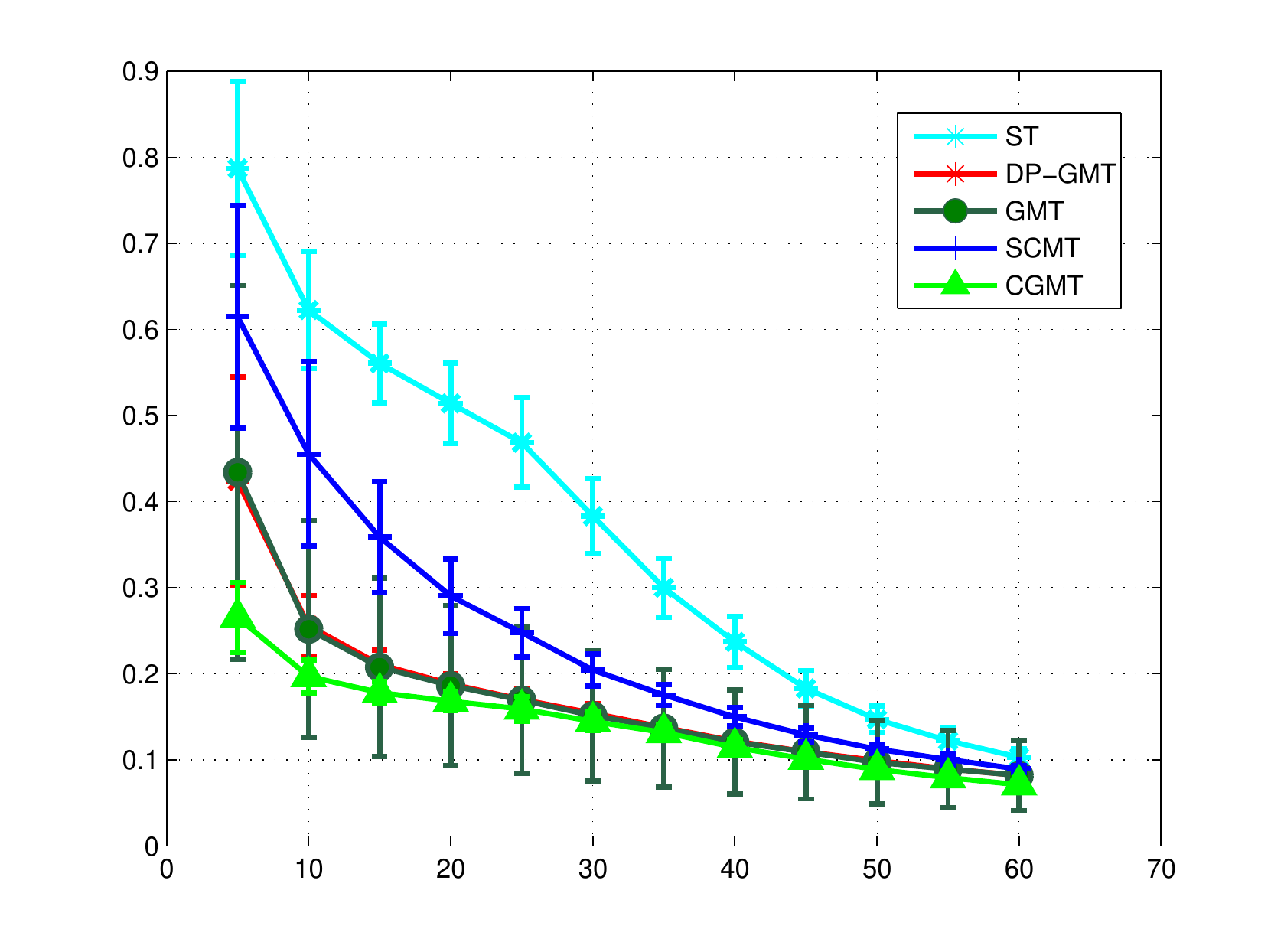}
        \caption{Simulated data: Performance comparison of single,
          multi-task, and grouped multi-task, and DP grouped multi-task as a
          function of the number of samples per task.}
        \label{fig:1b}
\end{figure}

Figure~\ref{fig:1a} shows a comparison of the algorithms for 50 random data
sets under the above setting when $N$ equals 5.
We see that
GMT with the correct model order $k=3$ almost always performs as well as its upper bound, illustrating that it recovers the
correct membership of each task. On only three data sets, our algorithm is trapped
in a local maximum yielding performance similar to SCMT and ST.
Figure~\ref{fig:1b} shows the RMSE for increasing values of $N$ for the same
experimental setup. From
the plot we can draw the conclusion that the proposed method works much
better than SCMT and ST when the number of samples is less than 30. As the
number of samples for each task increases, all methods are improving, but the
proposed method always outperforms SCMT and ST in our experiments. Finally,
all algorithms converge to almost the same performance level where
observations in each task are sufficient to recover the underlying function.
Finally,  Figure~\ref{fig:1b} also includes the performance of the DP-GMT
on the same data.
The
truncation level of the Dirichlet process is 10 and the
concentration parameter $\alpha$ is set to be 1.  As we can see
the DP-GMT is not distinguishable from the GMT (which has the correct $k$),
indicating that the model selection is successful in this example.



\subsection{Classification on Astrophysics data}
The concrete application motivating this research is the classification of stars into several
meaningful categories from the astronomy literature.
Classification is an important step within astrophysics research, as evidenced by published catalogs such as OGLE~\citep{udalski97ogle} and MACHO~\citep{Alcock1993,Faccioli2007}. However, the number of stars in such surveys is increasing dramatically. For example Pan-STARRS~\citep{Hodapp2004} and LSST~\citep{Starr2002} collect
data on the order of hundreds of billions of stars. Therefore, it is desirable to apply state-of-art machine learning techniques to enable automatic processing for astrophysics data classification.

\begin{figure}[t]
{\centering
\resizebox{1.5in}{!}{
\includegraphics{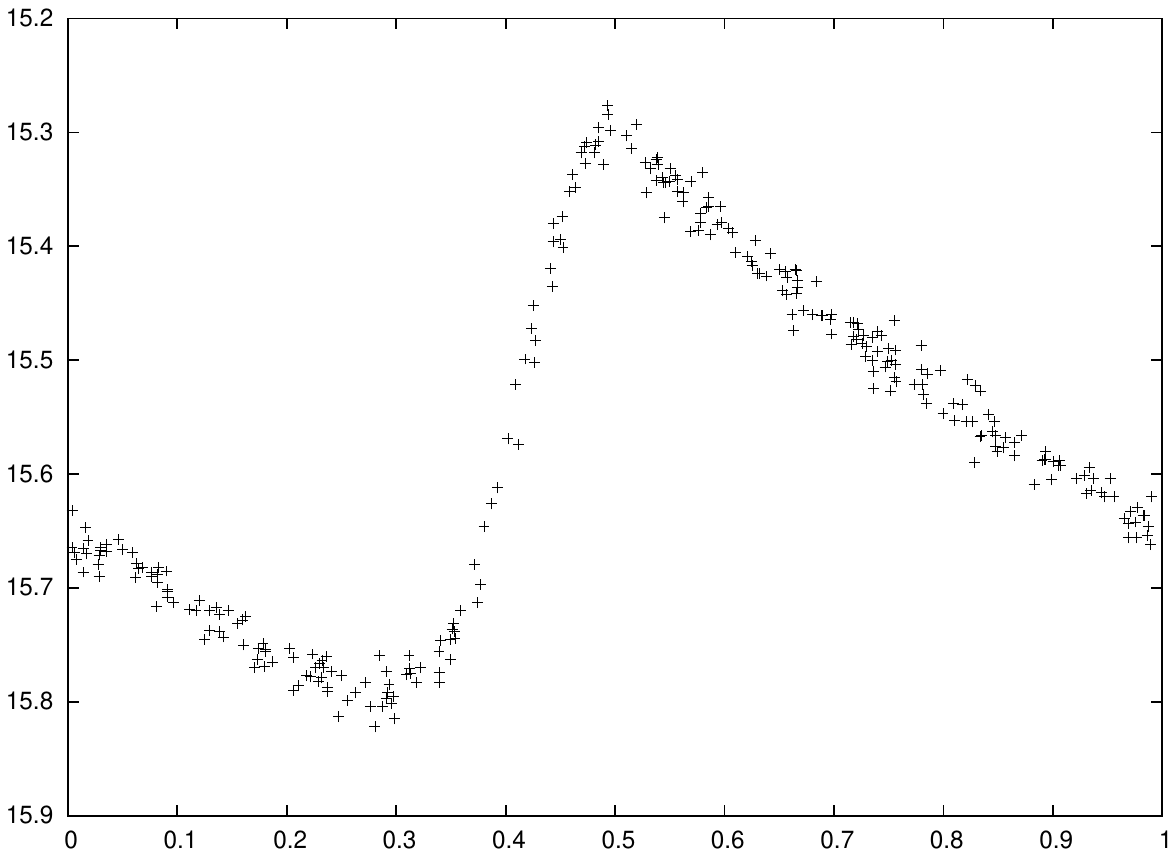}
}
\resizebox{1.5in}{!}{
\includegraphics{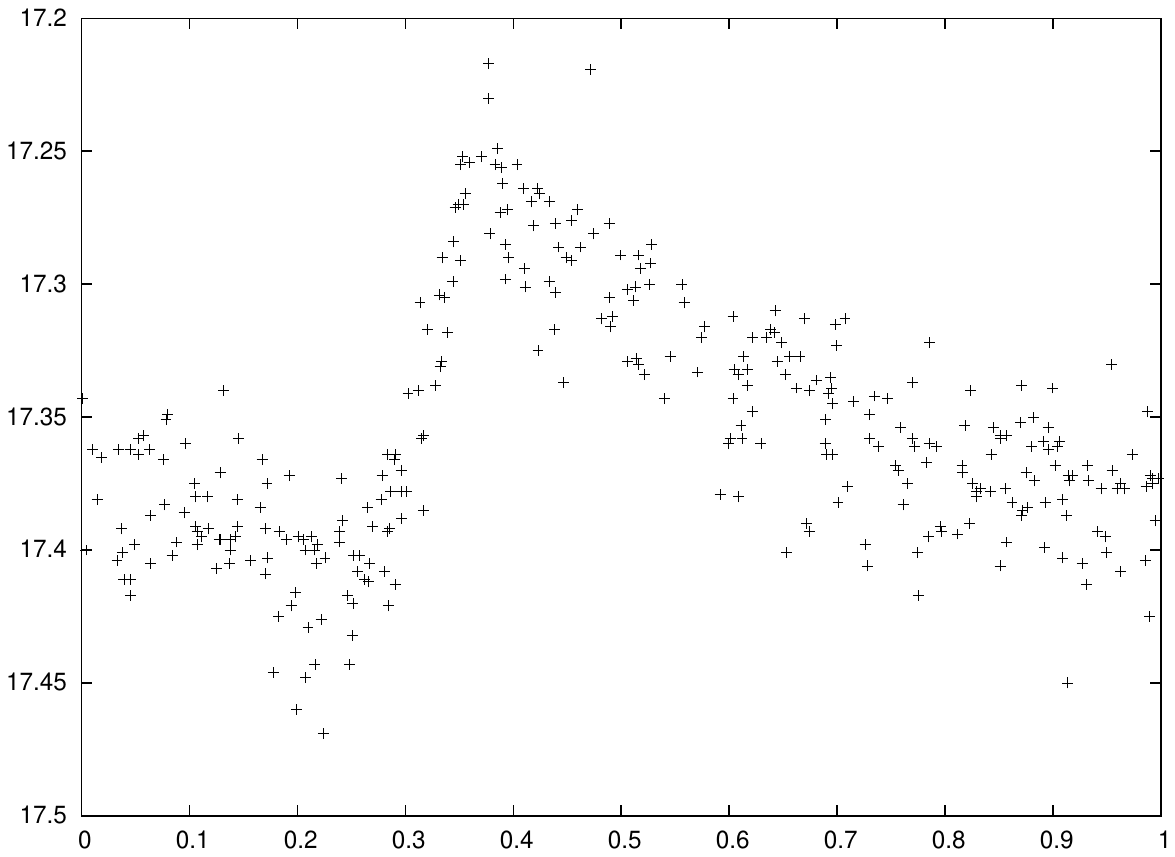}
}
\resizebox{1.5in}{!}{
\includegraphics{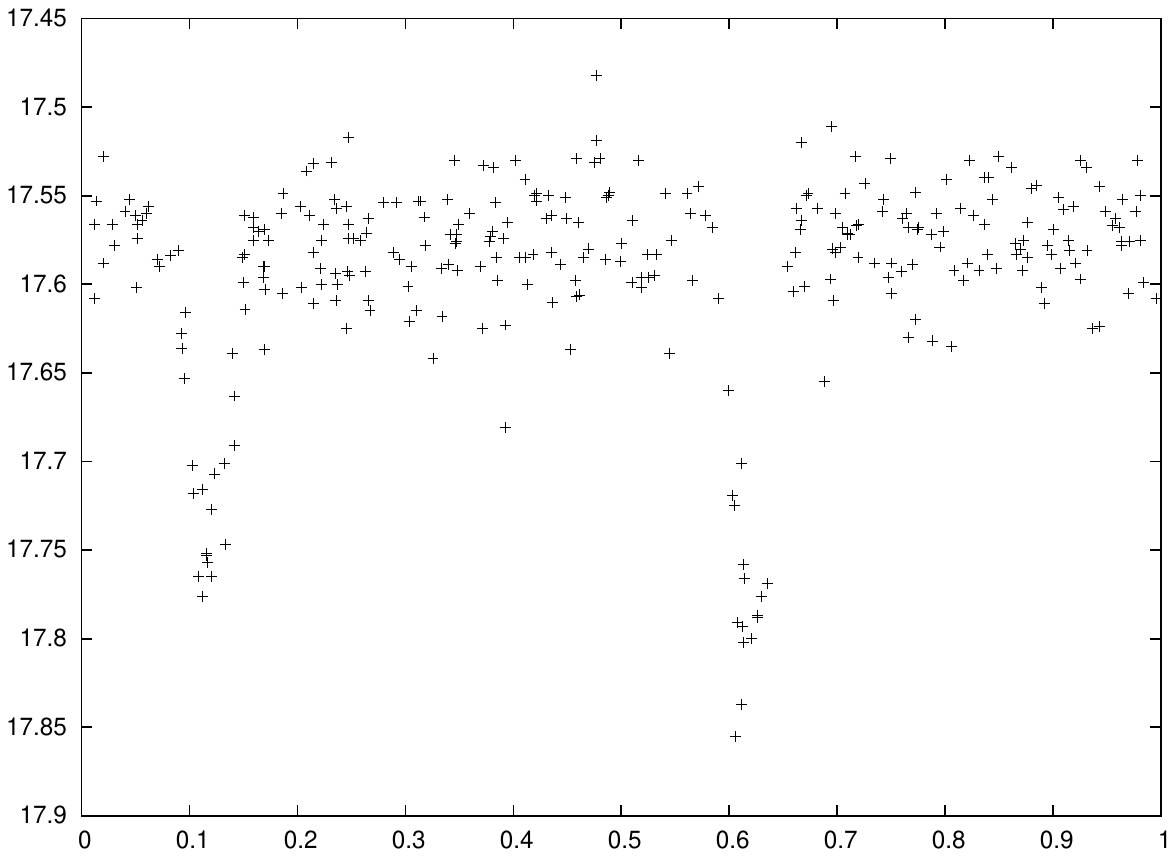}
}
\resizebox{1.5in}{!}{
\includegraphics{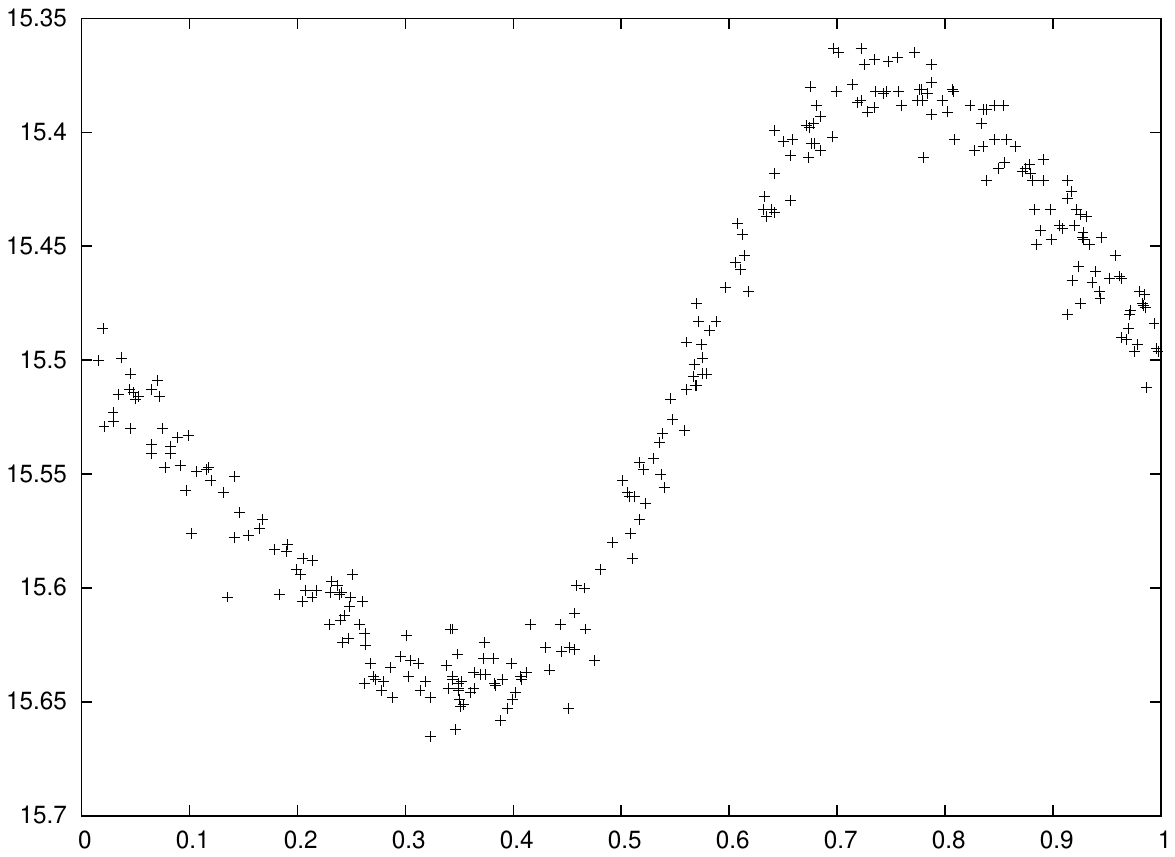}
}
\resizebox{1.5in}{!}{
\includegraphics{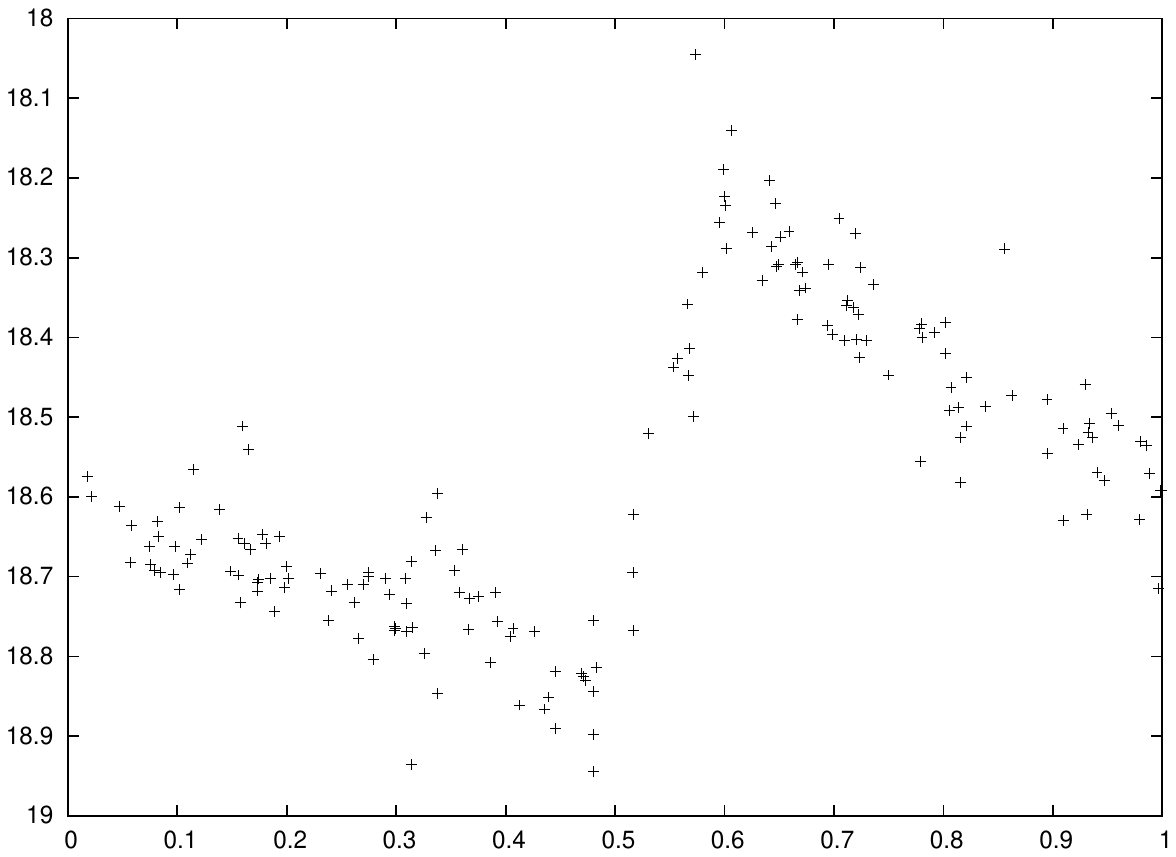}
}
\resizebox{1.5in}{!}{
\includegraphics{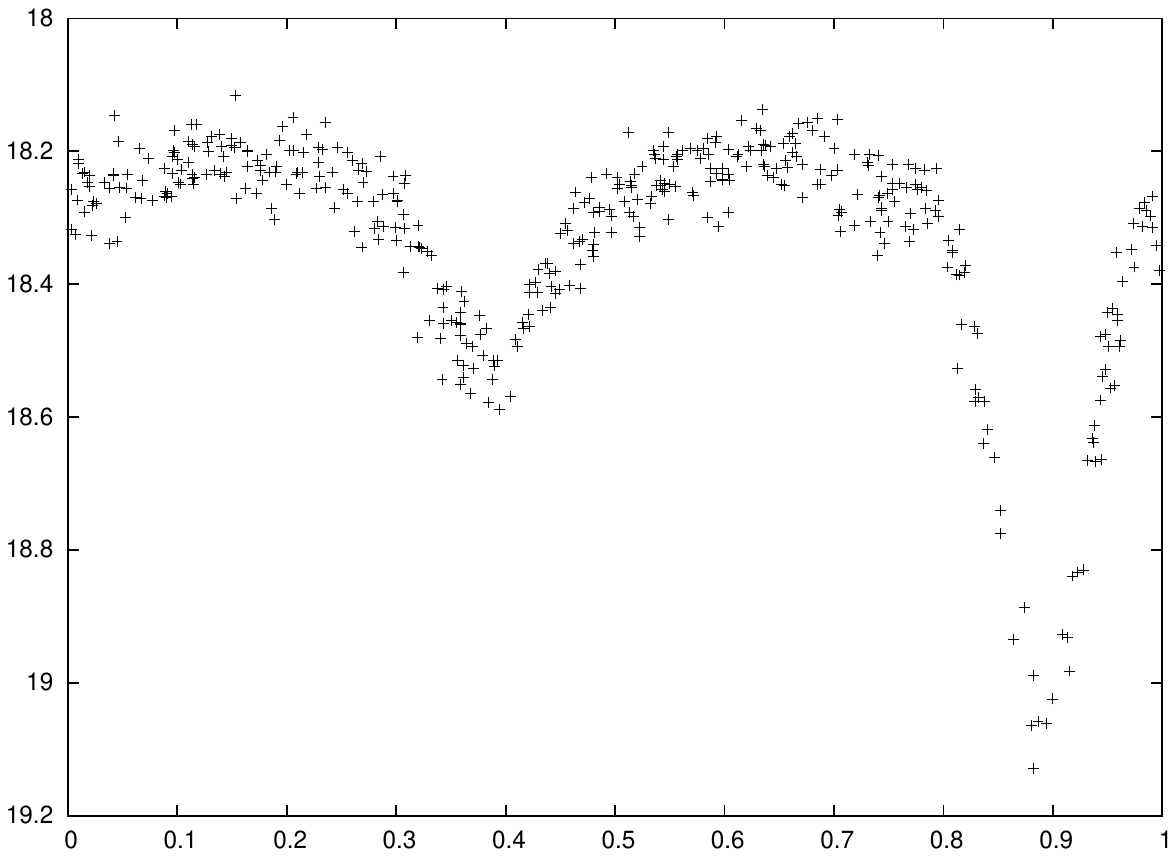}
}
\caption{Examples of light curves of periodic variable stars. Each column
  shows two stars of the same type. Left: Cepheid, middle: RR Lyrae,
  right: eclipsing binary.
}
\label{typical}
}
\end{figure}

The data from star surveys is normally represented by time series of brightness measurements,
based on which they are classified into categories. Stars whose behavior is periodic are especially of interest in such studies. Figure~\ref{typical} shows several examples
of such time series generated from the three major types of periodic variable
stars: Cepheid, RR Lyrae, and Eclipsing Binary. In our experiments only stars
of these classes are present in the data, and the period of each star is
given.

From Figure~\ref{typical}, it can be noticed that there are two main characteristics of this data set:
\begin{itemize}
  \item The time series are not phase aligned, meaning that the light curves in the same category share a similar shape but with some unknown shift.
  \item The time series are non-synchronously sampled and each light curve has different number of samples and sampling times.
\end{itemize}

We run our experiment on the OGLEII data
set~\citep{soszynski2003optical}. This data set consists of 14087 time series
from periodic variable stars with 3425 Cepheids, 3390 EBs and 7272 RRLs. We use
the time series measurements in the I band
~\citep{soszynski2003optical}. We perform several experiments with this data
set to explore the potential of the proposed method.
In previous work with this dataset \cite{wachman2009kernels} developed a kernel for
periodic time series and used it with \textsc{svm} to obtain good
classification performance. We use the results
of~\cite{wachman2009kernels} as our
baseline.\footnote{\cite{wachman2009kernels} used additional features,
in addition to time series itself, to improve the classification
performance. Here we focus on results using the time series
only. Extensions to add such features to our model are orthogonal to
the theme of the paper and we therefore leave them to future work in
the context of the application.}

\subsubsection{Classification using dense-sampled time series} In the first
experiment, the time series are smoothed using a simple average
filter, re-sampled to 50 points via linear-interpolation and
normalized to have mean 0 and standard deviation of 1. Therefore,
the time series are synchronously sampled in the pre-processing. We
compare our method to Gaussian mixture model (GMM) and 1-Nearest
Neighbor (1-NN). These two approaches are performed on the time
series processed by Universal phasing (UP), which uses the method
from~\cite{Protopapas06} to phase each time series according to the
sliding window on the time series with the maximum mean. We use a
sliding window size of $5\%$ of the number of original points; the
phasing takes place after the pre-processing explained above.
We learn a separate model for each class and
for each class the
model order for GMM and GMT is set to be 15.

\begin{table*}[t]
\centering
\begin{tabular}{||c||c|c|c|c||}
\hline
&  \textsc{up $+$ gmm} & \textsc{gmt} & \textsc{up $+$ 1-nn} & \textsc{k $+$ svm}\\
\hline
\textsc{Results} & $0.956\pm0.006$ & $0.952\pm 0.005$ & $0.865\pm 0.006$ & $0.947\pm 0.005$\\
\hline
\end{tabular}
\caption{Accuracies with standard deviations reported
on OGLEII dataset. }
\label{tab:fulacc}
\end{table*}

We run 10-fold cross-validation (CV) over the entire data set and
the results are shown in Table~\ref{tab:fulacc}. We see that when
the data is densely and synchronously sampled, the proposed
method performs similar to the GMM, and they both outperform the
kernel based results of~\cite{wachman2009kernels}. The similarity of the GMM and the proposed method under
these experimental conditions is not surprising. The reason is that
when the time series are synchronously sampled, aside from the
difference of phasing, finding the group effect functions is reduced
to estimating the mean vectors of the GMM. In addition, learning the
kernel in the non-parametric approach is equivalent to estimating
the covariance matrix of the GMM. More precisely, assuming all time
series are phased (that is, $t_j=0$ for all $j$), the following
results hold:

1. By placing a flat prior on the group effect function
$\f_s,s=1,\cdots,k$, or equivalently setting
$\|\f_s\|^2_{\mathcal{H}_0}=0$, Equation~(\ref{eqn:prob}) is reduced to
finding a vector $\mu_s\in\bbbn$ that minimizes
$\sum_j\gamma_{js}\|\ty^j - \mu_s\|^2$. Therefore, we obtain
$\f_s=\mu_s=\sum_j\gamma_{js}\ty_j/\sum_j\gamma_{js}$, which is
exactly the mean of the $s$th cluster during the iteration of
$\textsc{em}$ algorithm under the GMM setting.

2. The kernel $\mathbf{K}$ is learned in a non-parametric way. For
the GP regression model, we see that considering noisy observations is
essentially equivalent to considering non-noisy observations,
but slightly modifying the model by adding a diagonal term on the covariance
function for $\mathbf{f}_j$.
Therefore, instead of estimating $\mathbf{K}$ and
$\sigma^2$, it is convenient to put these two terms together,
forming $\widehat{\mathbf{K}}=\mathbf{K}+\sigma^2\mathbb{I}$.
In other words, we add a $\sigma^2$ term to the variance of $\mathbf{f}_j$
and remove it from $\y^j$ which becomes deterministic.
In this case, comparing to the derivation in
Equation~(\ref{eqn:14p})---(\ref{eqn:cjs}) we have
$\mathbf{f}_j = \y^j-\mathbf{\f}^j$
and $\mathbf{f}_j$ is determined given $z_j$.
Comparing to Equation~(\ref{eqn:fjconditional})
we have the posterior mean $\boldsymbol\mu_{js}^g =
\widehat{\mathbf{K}}\widehat{\mathbf{K}}^{-1}(\y^j - \mu_{s})=\y^j -
\mu_{s}$ and the posterior covariance matrix $\mathbf{C}_{j}^g$
vanishes. Applying these values in Equation~(\ref{eqn:nonp}) we get
$\widehat{\mathbf{K}} =
\frac{1}{M}\sum_j\sum_s\gamma_{js}(\y^j-\mu_s)(\y^j-\mu_s)^\T$. In
the standard $\textsc{em}$ algorithm for the GMM, this is equal to
the estimated covariance matrix when all $k$ clusters are assumed to
have the same variance.

Accordingly, when time series are synchronously sampled, the
proposed model can be viewed as an extension of the Phased
K-means~\citep{rebbapragada2009finding}. The Phased K-means
(PKmeans) re-phases the time series before the similarity
calculation and updates the centroids using the phased time series.
Therefore, with shared covariance matrix, our model is a
shift-invariant (Phased) GMM and the corresponding learning process
is a Phased \textsc{em}
 algorithm where each time series is re-phased in the \textbf{E} step. In
 experiments presented below we use Phased GMM directly in the feature space and generalize it so that each class has a separate
 covariance matrix.

\begin{figure}
    \centering
        \includegraphics[width=10cm]{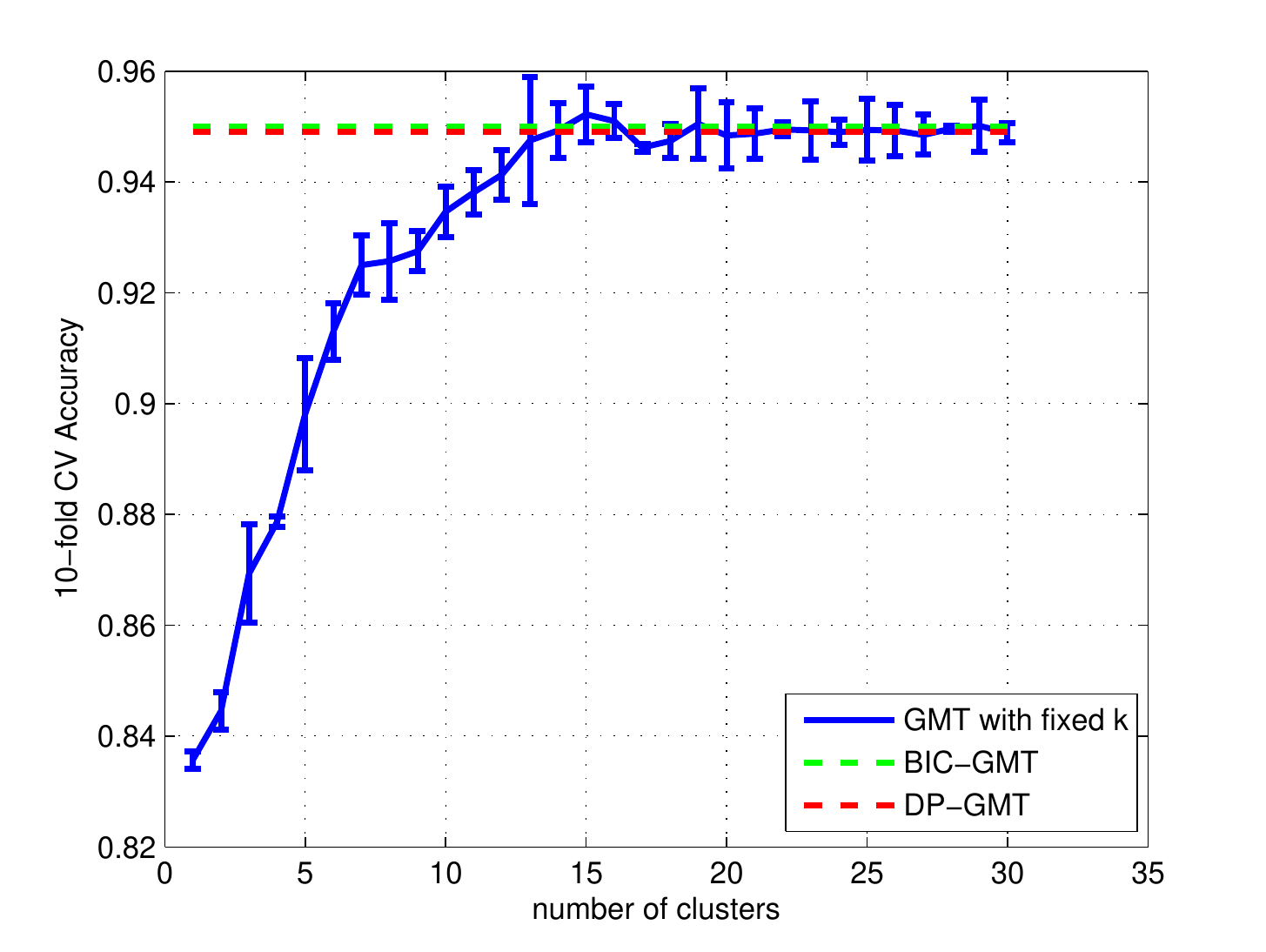}
        \caption{OGLEII data: Comparison of model selection methods using
          dense sampled data. The plot shows the performance of
GMT with varying $k$, BIC for the GMT model, and DP-GMT. For visual clarity
we only include the standard deviations on the GMT plot. }
        \label{fig:dense2}
    \end{figure}

We use the same experimental data to
investigate the performance of the DP-GMT where the
truncation level is set to be 30 and the
concentration parameter $\alpha$ of the DP is set to be 1.
The results are shown in Figure~\ref{fig:dense2} and Table~\ref{tab:fulacc2}
where BIC-GMT means that
the model order is chosen by BIC  where the optimal $k$ is chosen from 1 to 30. The poor
performance of SCMT shows that a single center is not sufficient for
this data.
As evident from the graph
the DP-GMT is not distinguishable from the BIC-GMT.
The advantage of the DP model is that this equivalent performance is achieved
with much reduced computational cost because the BIC procedure must learn
many models and choose among them whereas the DP learns a single model.

\begin{table*}[t]
\centering
\begin{tabular}{||c||c|c|c|c||}
\hline
&  \textsc{scmt} & \textsc{gmt} & \textsc{dp-gmt} & \textsc{bic-gmt}\\
\hline
\textsc{Results}  & $0.874 \pm 0.008$ & $0.952\pm 0.005$ & $0.949\pm 0.005$ & $0.950\pm 0.002$\\
\hline
\end{tabular}
\caption{Accuracies with standard deviations reported
on OGLEII dataset. }
\label{tab:fulacc2}
\end{table*}

\subsubsection{Classification using sparse-sampled time series} The OGLEII data set is in some sense a ``nice'' subset of the data from its corresponding star survey. 
Stars with small number of samples are often removed in pre-processing steps. For example, \cite{wachman2009thesis} developed full system to process the MACHO catalog and applied the kernel method to classify stars. In its pipeline, part of the preprocessing rejected 3.6 million light curves of the approximate 25 million because of insufficient number of observations. The proposed method potentially provides a way to include these instances in the classification process.  In the second experiment,
we demonstrate the performance of the proposed method on times
series with sparse samples. Similar to the synthetic data, we
started from sub-sampled versions of the original time series to
simulate the condition that we would encounter in further star
surveys.\footnote{For the proposed method, we clip the samples to a
  fine grid of 200 equally spaced time points on $[0,1]$, which is also the set of allowed time shifts. This avoids having a very high dimensional $\breve{\x}$,
  e.g. over 18000 for OGLEII, which is not feasible for any kernel based
  regression method that relies on solving linear systems.}
As in the previous experiment, each time series is universally
phased, normalized and linearly-interpolated to length 50 to be
plugged into GMM and 1-NN as well as the phased GMM mentioned above.
The RBF kernel is used for the proposed
method and we use model order 15 as above. Moreover, the
performance for PKmeans is also presented, where the classification
step is as follows: we learn the PKmeans model with $k=15$ for each
class and then the label of a new example is assigned to be the same
as its closest centroid's label. PKmeans is also restarted 5 times
and the best clustering is used for classification.

    \begin{figure}
    \centering
        \includegraphics[width=10cm]{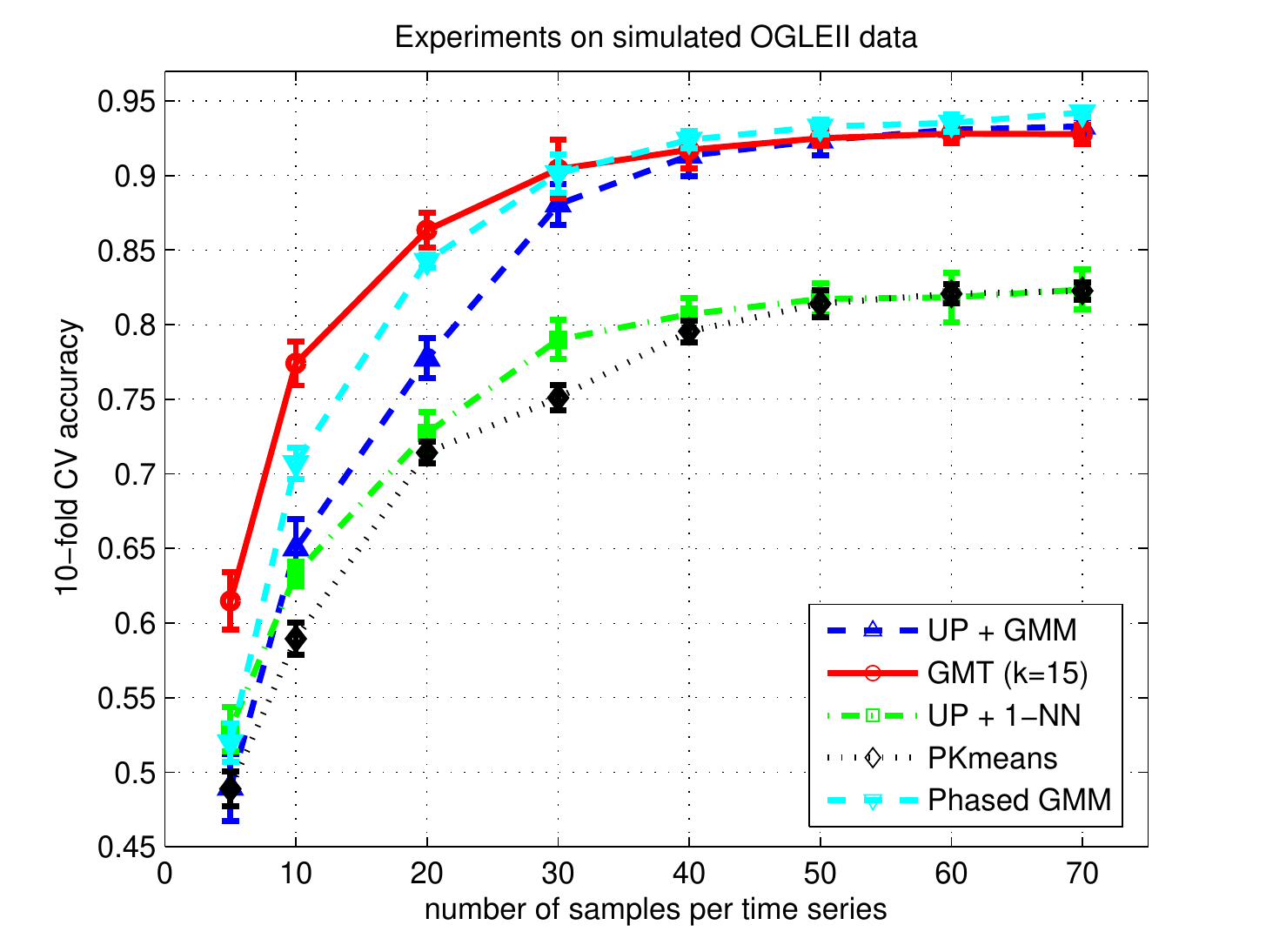}
        \caption{OGLEII data: Comparison of algorithms with sparsely sampled data}
        \label{fig:sub1}
    \end{figure}
    \begin{figure}
    \centering
        \includegraphics[width=10cm]{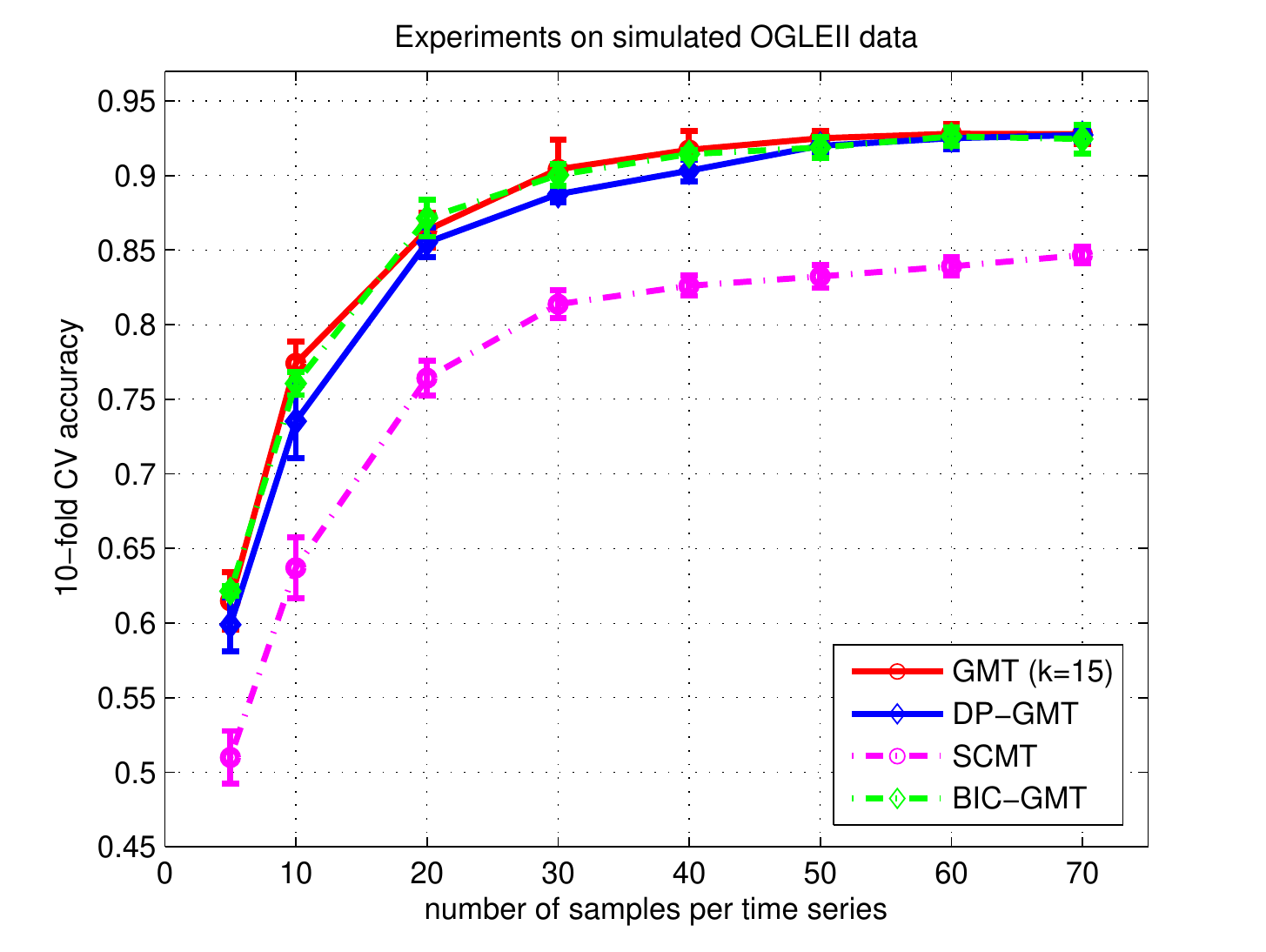}
        \caption{OGLEII data: Comparison of algorithms with sparsely sampled data}
        \label{fig:sub2}
    \end{figure}

The results are shown in Figure~\ref{fig:sub1}. As can be easily
observed, when each time series has sparse samples (i.e., number of
samples per task is less than 30), the proposed method has a
significant advantage over the other methods. As the number of
samples per task increases, the proposed method improves fast and
performs close to its optimal performance given by previous
experiment.
Three additional aspects that call for discussion can be seen in the
figure. First, note that for all three methods, the performance with
dense data is lower than results in Table~\ref{tab:fulacc}.
This can be explained
by fact that the data set obtained by the interpolation of the
sub-sampled measurements contains less information than that
interpolated from the original measurements. Second, notice that the
Phased \textsc{em} algorithm always outperforms the GMM plus UP
demonstrating that re-phasing the time series inside the \textsc{em}
algorithm improves the results. Third, when the number of samples
increases, the performance of the Phased \textsc{em} gradually
catches up and becomes better than the proposed method when each
task has more than 50 samples. GMM plus universal phasing (UP) also
achieves better performance when time series are densely sampled.
One reason for the performance difference is the difference in the
way the kernel is estimated. In Figure~\ref{fig:sub1} GMT uses the
parametric form of the kernel which is less expressive than getting
precise estimates for every $\mathcal{K}(t_1,t_2)$. The GMM uses the
non-parametric form which, given sufficient data, can lead to better
estimates. A second reason can be attributed to the sharing of the
covariance function in our model where the GMM and the Phased GMM do
not apply this constraint.

Finally, we use the same experimental setting to
compare the performance of various mode selection models.
The results are shown in Figure~\ref{fig:sub2}.
The performance of BIC is not distinguishable from the optimal $k$ selected
in hindsight. The performance of DP is slightly lower but it comes close to
these models.

To summarize, we conclude from the experiments with astronomy data that Phased
\textsc{em} is appropriate with densely sampled data but that the
GMT and its variants should be used when data is sparsely and
non-synchronously sampled. In addition BIC coupled with GMT performs
excellent model selection and DP does almost as well with a much reduced
computational complexity.

\subsubsection{Class discovery: }

We show the potential of our model for class discovery by running two version
of the GMT model on the joint data set of the three classes (not using the
labels).  Then, each cluster is labeled according to the majority class of
the instances that belong to the center. For a new test point, we determine
which cluster it belongs to via the MAP probability and its label is given by
the cluster that it is assigned to. We run 10 trials with different random
initializations.
In accordance with previous experiments that used 15 components per class
we run GMT with model order of 45. We also run DP-GMT
with a truncation level set to 90.
The GMT obtains accuracy and standard deviation of $[0.895, 0.010]$
and the DP models obtains accuracy and standard deviation of
$[0.925, 0.013]$.
Note that it is hard to compare between the results because of the different
model orders used. Rather than focus on the difference, the striking point is
that we obtain almost pure clusters without using any label information.
Given the size
of the data set and the relatively small number of clusters this is
a significant indication of the potential for class discovery in
astrophysics.


\section{Related Work}

Classification of time series has attracted an increasing amount of interest in
recent years due to its wide range of potential applications, for example ECG
diagnosis~\citep{wei2006sst}, EEG diagnosis~\citep{lu2008rkh}, and Speech
Recognition~\citep{povinelli2004tsc}.  Common methods choose some feature
based representation or distance function
for the time series (for example the sampled time
points, or Fourier or wavelet coefficients as features
and dynamic time warping for distance function) and then apply some existing
classification method
~\citep{osowski2004svm,ding2008qam}.  Our approach falls into another
category, that is, model-based classification where the time series are
assumed to be generated by a probabilistic model and examples are classified
using maximum likelihood or MAP estimates.  A family of such models, closely
related to the GMT, is discussed in detail below.  Another common approach
uses Hidden Markov models as a probabilistic model for sequence
classification, and this has been applied to time series as well
\citep{kim2006segmental}.


Learning Gaussian processes from multiple tasks has previously been
investigated in the the hierarchical Bayesian framework, where a group of
related tasks are assumed to share the same prior. Under this assumption,
training points across all tasks are utilized to learn a better covariance
function via the \textsc{em}
algorithm~\citep{yu2005learning,schwaighofer2005lgp}. In addition,
\cite{lu2008rkh} extended the work of~\cite{schwaighofer2005lgp} to a
non-parametric mixed-effect model where each task can have its own random
effect. Our model is based on the same algorithmic approach where the values
of the function for each task at its corresponding points
(i.e. $\{\mathbf{f}_j\}$ in our model) are considered as hidden
variables. Furthermore, the proposed model is a natural generalization
of~\cite{schwaighofer2005lgp} where the fixed-effect function is sampled from
a mixture of regression functions each of which is a realization of a common
Gaussian process.  Along a different dimension, our model differs from the
infinite mixtures of Gaussian processes model for clustering
\citep{jackson2007bus} in two aspects: first, instead of using zero mean
Gaussian process, we allow the mean functions to be sampled from another Gaussian process; second,
the individual variation in our model serves as the covariance function in their model but all mixture components
share the same kernel.

Although having a similar name,
the Gaussian process {\em mixture of experts} model
focuses mainly on the issues of non-stationarity in
regression~\citep{rasmussen2002infinite,tresp2001mixtures}. By dividing the
input space into several (even infinite) regions via a gating network, the
Gaussian process mixture of expert model allows different Gaussian processes to make
predictions for different regions.

In terms of the clustering aspect, our work is most closely related to the
so-called \emph{mixture of regressions}~\citep{gaffney2005joint,gaffney2003curve,gaffney2004probabilistic,gaffney1999trajectory}. The
name comes from the fact that these approaches substitute component density
models with conditional regression density models in the framework of
standard mixture model. For phased time series, \cite{gaffney1999trajectory}
first proposed the regression-based mixture model where they used Polynomial
and Kernel regression models for the mean curves. Further,
\cite{gaffney2003curve} integrated the linear random effects models with
mixtures of regression functions. In their model, each time series is sampled
by a parametric regression model whose parameters are generated from a
Gaussian distribution.
To incorporate
the time shifts, \cite{chudova2003translation} proposed a shift-invariant
Gaussian mixture model for multidimensional time series. They constrained the
covariance matrices to be diagonal to handle the non-synchronous
case. They also treated time shifts as hidden variables and derived the
\textsc{em} algorithm under full Bayesian settings, i.e. where each parameter
has a prior distribution. Furthermore, \cite{gaffney2005joint} developed a
generative model for misaligned curves in a more general setting. Their joint
clustering-alignment model also assumes a normal parametric regression model
for the cluster labels, and Gaussian priors on the hidden transformation
variables which consist of shifting and scaling in both the time and
magnitude. Our model extends the work of~\cite{gaffney2003curve} to admit
non-parametric Bayesian regression mixture models and at the same time handle
the non-phased time series. If the group effects are assumed to have a flat
prior, our model differs from~\cite{chudova2003translation} in the following
two aspects in addition to the difference of Bayesian treatment. First, our
model does not include the time shifts as hidden variables but instead
estimates them as parameters. Second, we can handle shared full covariance
matrix instead of diagonal ones by using a parametric form of the
kernel. On the other hand, given the time grid $\breve{\x}$, we can design
the kernel
for individual variations as $\mathcal{K}(\breve{x}_i,\breve{x}_j) =
a_i\delta_{ij}(\breve{x}_i,\breve{x}_j), i,j=1,\cdots,\bbbn$. Using this
choice, our model is the same as~\cite{chudova2003translation} with shared
diagonal covariance matrix. In summary, our model allows a more flexible
structure of the covariance matrix that can treat synchronized and
non-synchronized time series in a unified framework, but at the same time it
is constrained to have the same covariance matrix across all clusters.

\section{Conclusion}

We developed a novel Bayesian nonparametric multi-task learning model (GMT)
where each task is modeled as a sum of a group-specific function and an
individual task function with a Gaussian process prior. We also extended the
model such that the number of groups is not bounded using a Dirichlet process
mixture model (DP-GMT).
We
derive efficient \textsc{em} and variational \textsc{em} algorithms to learn
the parameters of the models and demonstrated their effectiveness using
experiments in regression, classification and class discovery. Our models are
particularly useful for sparsely and non-synchronously sampled time series
data, and model selection can be effectively performed with these models.

There are several natural directions for future work. For application in the
astronomy context it is important to consider all steps of processing and
classification of a new sky survey so as to provide an end to end
system. Therefore, two important issues to be addressed in future work
include incorporating the period estimation phase into the method and
developing an appropriate method for abstention in the classification
step. It would also be interesting to develop a corresponding discriminative
model extending \cite{xue2007multiBB} to the GP context. Finally, one of the
drawbacks of the GP based methods is the computational complexity which is
too high for large scale problems. For example, in the experiments on
sparse OGLEII data, we had to resample the data on a fine grid to avoid
performing Cholesky decomposition for high dimensional matrices.
Therefore, an important direction for future work
is to find non-trivial
sparse GP approximations that yield good performance with the GMT model.


\acks{This research was partly supported by NSF grant IIS-0803409.
The experiments in this paper were performed on the
Odyssey cluster supported by the FAS Research Computing Group at
Harvard and the Tufts Linux Research Cluster supported by Tufts UIT Research Computing. }

\bibliographystyle{natbib}
\bibliography{npb}

\end{document}